\documentclass[final,1p,times]{elsarticle}

\usepackage{graphics} 
\usepackage{epsfig} 
\usepackage{times} 
\usepackage{amsmath} 
\usepackage{amssymb}  
\usepackage[table]{xcolor}
\usepackage[ruled,linesnumbered, noend]{algorithm2e}

\usepackage{comment}
\usepackage{multirow}
\usepackage{graphicx}
\usepackage{wrapfig}
\usepackage{subfigure}
\usepackage{balance}
\usepackage{wrapfig}
\usepackage{booktabs}  
\usepackage{float}

\newcommand{\ve}[1]{\mathbf{#1}} 
\newcommand{\tve}[1]{\tilde{\mathbf{#1}}} 
\newcommand{\bve}[1]{\bar{\mathbf{#1}}} 
\newcommand{\hve}[1]{\hat{\mathbf{#1}}} 

\usepackage{soul, color}
\usepackage{colortbl}
\definecolor{Gray}{gray}{0.9}

\usepackage{pifont}

\newcommand{\hq}[0]{HQ-Net } 
\newcommand{\gv}[0]{GViT } 

\journal{ }

\begin{document}

\begin{frontmatter}



\title{Ultra-Range Gesture Recognition using a Web-Camera in Human-Robot Interaction}


\author[label1]{Eran Bamani} 
\author[label1]{Eden Nissinman} 
\author[label1]{Inbar Meir} 
\author[label1]{Lisa Koenigsberg} 
\author[label1]{Avishai Sintov\corref{cor1}}
\affiliation[label1]{organization={School of Mechanical Engineering, Tel-Aviv University},
            addressline={Haim Levanon St.},
            city={Tel-Aviv},
            postcode={6997801},
            country={Israel}}
\ead{sintov1@tauex.tau.ac.il}
\cortext[cor1]{Corresponding Author.}
%





\begin{abstract}

Hand gestures play a significant role in human interactions where non-verbal intentions, thoughts and commands are conveyed. In Human-Robot Interaction (HRI), hand gestures offer a similar and efficient medium for conveying clear and rapid directives to a robotic agent. However, state-of-the-art vision-based methods for gesture recognition have been shown to be effective only up to a user-camera distance of seven meters. Such a short distance range limits practical HRI with, for example, service robots, search and rescue robots and drones. In this work, we address the Ultra-Range Gesture Recognition (URGR) problem by aiming for a recognition distance of up to 25 meters and in the context of HRI. We propose the URGR framework, a novel deep-learning, using solely a simple RGB camera. Gesture inference is based on a single image. First, a novel super-resolution model termed High-Quality Network (HQ-Net) uses a set of self-attention and convolutional layers to enhance the low-resolution image of the user. Then, we propose a novel URGR classifier termed Graph Vision Transformer (\gv\!\!) which takes the enhanced image as input. \gv combines the benefits of a Graph Convolutional Network (GCN) and a modified Vision Transformer (ViT). Evaluation of the proposed framework over diverse test data yields a high recognition rate of 98.1\%. The framework has also exhibited superior performance compared to human recognition in ultra-range distances. With the framework, we analyze and demonstrate the performance of an autonomous quadruped robot directed by human gestures in complex ultra-range indoor and outdoor environments, acquiring 96\% recognition rate on average.

\end{abstract}




\end{frontmatter}

\section{Introduction}
\label{sec:introduction}

Gestures are an imperative medium to complement verbal communication between humans \cite{krauss1996nonverbal,bernardis2006speech,cook2018enhancing}. From simple hand gestures to complex body movements, gestures can convey rapid and nuanced information that is essential for effective communication. Often, gestures entirely replace speech when the delivered message is short and simple \cite{goldin1999role}, e.g., thumbs-up for approval or beckoning for an invitation to approach. Also, a gesture is the simplest and most immediate communication when the source and target participants are far from each other. In such a case, speech may not be clearly conveyed. Hence, if a line-of-sight between the participants exists, a gesture can transfer quick and clearer messages or commands. 
\begin{figure}
    \centering
    \includegraphics[width=0.55\linewidth]{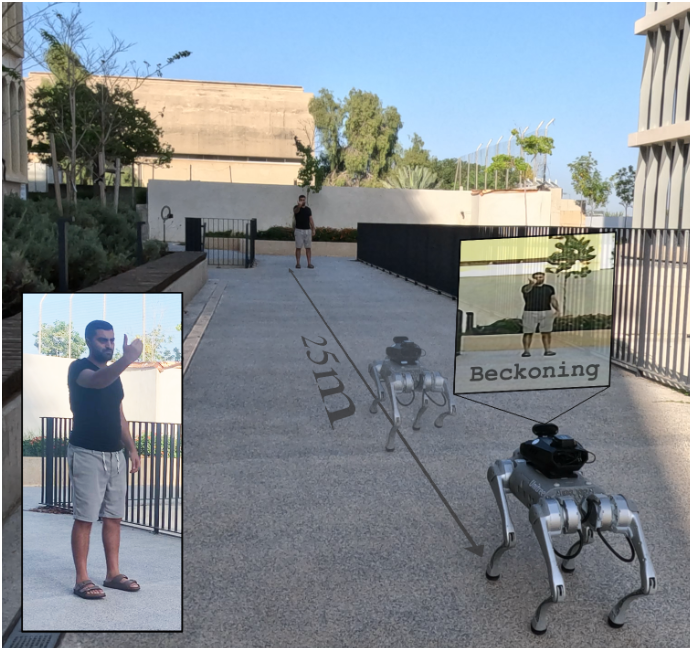}
    \caption{A robot recognizing a directive gesture from a user located 25 meters away by solely using an RGB camera. Upon recognizing, for instance, a beckoning gesture, the robot will move toward the user.}
    \label{fig:withGo1}
\end{figure}
\begin{figure*}
    \centering
    \includegraphics[width=\linewidth]{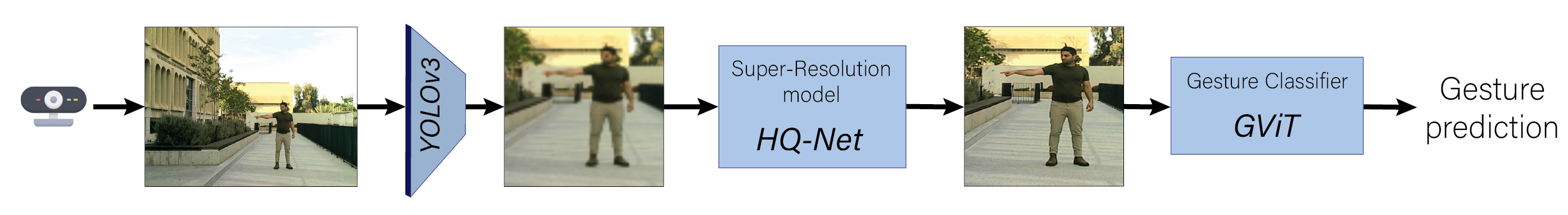}
    \caption{Illustration scheme of the proposed URGR framework. The user in the image is detected with YOLOv3 followed by cropping the background. Since the user is in low quality due to the large distance from the camera, HQ-Net is a proposed super-resolution method that enhances the quality of the cropped image. Then, a classification model termed \gv outputs the predicted gesture.}
    \label{fig:scheme}
\end{figure*}

Similar to humans, in Human-Robot Interaction (HRI), a user may wish to convey a directive message to a robot using the same medium. For instance, the user may wish to exhibit a desired motion direction for the robot through a pointing gesture \cite{bamani2023recognition}. Therefore, efficient gesture recognition in HRI is indispensable in order to acquire natural conveyance of commands and information to robots. Since the leading approach for gesture recognition is visual perception \cite{nickel2007visual,wachs2011vision,xia2019vision,weng2022development,Gao2022}, an important aspect is the distance range between the gesturing user and the camera on an agent. Previous work has commonly considered short- and long-range gesture recognition roughly limiting both up to 1 meter and 4-7 meters, respectively \cite{Liu2022}. However, we argue that long-range gesture recognition can be increased further away. With a longer effective range, gesture communication with a robot is more practical and the robot is not bound to the close proximity of the user during tasks. A longer recognition range extends the operational workspace of the robot, enabling effective navigation in larger areas. This also benefits individuals with limited mobility and expands accessibility while also playing a vital role in diverse applications such as surveillance, search and rescue, and outdoor robotics. Therefore, we define the \textit{Ultra-Range Gesture Recognition} (URGR) problem where effective recognition is required up to a distance of 25 meters. This advancement enhances a robots' ability to assist and engage with humans across various contexts, thereby broadening the landscape of human-robot collaboration.

Short-range (or close-range) gesturing can generally be used when working in front of a computer \cite{oudah2020hand,An2022}, with an infotainment system within a car \cite{AlbaCastro2014,Buddhikot2018} or in Virtual Reality (VR) \cite{Deller2006}. Close-range approaches for gesture recognition are, in most cases, not feasible for HRI with mobile systems. On the other hand, long-range gesturing may include indoor service robots such as cleaning robots, robotic servants and medical robots. Also, applications are considered in smart homes, TV control and video games \cite{Zulpukharkyzy2021}. While numerous approaches have been proposed for long-range gesture recognition with high success rates, these were only demonstrated indoors, with a relatively structured background, and the majority were effective only up to 4-5 meters \cite{Mazhar2018,Chang2019}. Some of the approaches use an RGB-Depth (RGB-D) camera \cite{nickel2007visual,Iengo2014,Nguyen2019}. Depth cameras are limited to short-range in indoor environments and work poorly outdoors. Furthermore, they require an additional hardware setup and limit the generality of the method. With RGB cameras, various gesture recognition packages are available open-source (e.g., SAM-SLR \cite{Jiang2021}, MediaPipe Gesture Recognizer and OpenPose \cite{Cao2019}) and work quite well in short- and long-range. However, these were tested in a preliminary study yielding poor results beyond 4 meters.

While long-range gesture recognition can offer a high recognition rate, it may not be practical in actual HRI tasks due to a limited range. The farthest distance reported for visual-based gesture recognition is 7 meters. One work reaching such a distance involved an attention-based model able to recognize short- and long-range hand gestures \cite{zhou2021long}. Similarly, a model based on convolutional and feature aggregation pyramid networks achieved effective feature extraction for gesture detection and recognition \cite{Liang2024}. However, these approaches were not demonstrated in ultra-range or outdoor environments. To the best of the authors' knowledge, no work has addressed the gesture recognition problem at distances farther than 7 meters. 

In this paper, we address and explore the URGR problem and aim for an effective distance of up to 25 meters solely based on an RGB camera. Hence, a user can effectively direct a robot from a large distance (Figure \ref{fig:withGo1}). We propose a data-based framework which does not require depth information but solely a simple web-camera. Hence, the approach is accessible and cost-effective. While temporal inference may be useful, high-frequency video streams may not always be available and processing such a video may introduce computational overhead. Therefore, we focus on the challenge of recognition from static gestures in images. Future work may build upon this to include temporal information.

The main challenge in URGR is the low resolution of the image after cropping out the background and focusing on the gesturing user. The resulted image is usually distorted with compromised details of the hand due to the extended distance. To overcome the low-resolution challenge, Super Resolution (SR) is a common method to effectively enhance the resolution of an image and enable more precise object recognition \cite{WangZ2021}. Algorithms, such as ESRGAN \cite{Wang2019} and Real-ESRGAN \cite{Wang2021}, utilize a Generative Adversarial Network (GAN) in order to determine whether a reconstructed image is realistic. Hence, the GAN guides the recovery of the image toward reconstructing details. However, these algorithms may prioritize low-resolution information at the expense of high-frequency information, leading to over-smoothing and distortion of indistinguishable features (e.g., fingers). While these models may focus on static or prominent elements of the image such as windows and bricks, they may not accurately capture intricate details of the human body in various postures. Therefore, in this work, we propose a novel SR model termed \textit{High-Quality Network} (HQ-Net) which is based on a scheme of filters, self-attention and convolutional layers. HQ-Net may render the use of expensive, high-resolution cameras unnecessary, and a simple web camera is sufficient. \hq is validated through a comparison to state-of-the-art in SR with a focus on the URGR problem.

In our proposed URGR framework, visualized in Figure \ref{fig:scheme}, an image taken by the camera is pre-processed to focus on the user. Quality improvement is followed using the \hq model. Then, the improved image is fed into a classifier termed \textit{Graph-Vision Transformer} (\gv\!\!). The \gv model leverages the benefits of the Graph Convolutional Network (GCN) \cite{Ullah2019} and Vision Transformers (ViT) \cite{dosovitskiy2020image}. ViT is modified to receive embedded information from the graph convolutional layers of the GCN. The combination of GCN and ViT in \gv enables it to capture local and global dependencies in the image. The \gv outputs the probability distribution over all gesture classes. The model was integrated into a robotic system which executes actions corresponding to recognized gestures. To summarize, the contributions of this work are as follows:
\begin{itemize}
    \item A novel SR model, HQ-Net, is proposed for significantly enhancing the quality of low-resolution images. In addition to significantly improving the success rate in URGR, it may be used for traditional applications of SR such as image processing, security and medical imaging.
    \item During the training of the \hq model, we offer an image degradation process specifically designated for ultra-distance objects. The process is validated by comparing it to other methods over URGR test cases. 
    \item Unlike prior work, the proposed GViT model is the first to recognize gestures in ultra-range up to a distance of 25 meters between the camera and the user. The model is able to work in complex indoor and outdoor environments and is also evaluated in various edge cases such as with low lighting conditions and occlusions. Also, the model is shown to perform better than the performance of human recognition.
    \item The proposed approach is demonstrated and evaluated in an HRI scenario with a mobile robot.
    \item Trained models and datasets are available open-source for the benefit of the community\footnote{To be available upon acceptance for publication. Images will be modified to protect the privacy of the participants.}.
\end{itemize}

The proposed approach may be used in the same applications of long-distance, as described above, but with an extended and more practical range. Furthermore, URGR may be required for directing search and rescue robots, drones and other service robots. Space exploration, police work and entertainment can also gain from URGR. The work may also be adapted in future work to various other object recognition tasks with an ultra-range capability.

\section{Related Work}
\label{sec:relatedwork}

\subsection{Gesture Recognition}
Gesture recognition is a long-standing research field in computer vision \cite{Brethes2004}. The common approaches are based on capturing hand poses or movements using either an RGB camera, an RGB-D camera or some wearable device. RGB-D cameras are a popular tool in gesture recognition since they add spatial information to the observed RGB image. Hence, they are able to provide more accurate information regarding the exhibited gesture \cite{ma2018kinect,zhu2017multimodal,nakamura2023deepoint}. Despite enabling easier access to gesture features, depth cameras work poorly outdoors and are usually limited to short to long ranges in indoor environments. In addition, they require an additional hardware setup and limit the generality of the method \cite{Jirak2020}. To address these limitations, researchers have turned to recognizing human gestures using RGB cameras which have lower cost and are more accessible. Early approaches handcrafted methods for extracting features of hand pose for further recognition with traditional classifiers such as Support Vector Machine (SVM) \cite{Huang2009} or k-Nearest Neighbors (kNN) \cite{Ziaie2009}. However, such approaches can usually handle short-range recognition and may have difficulties in unstructured backgrounds. Hence, more recent work focused on deep-learning approaches \cite{An2022, zhou2021long,Liang2024}. For instance, an approach combined Convolutional Neural Network (CNN) with Long Short-Term Memory (LSTM) for dynamic gesture recognition in short-range with a single camera \cite{eleni2015}. Another work utilized a single moving camera to capture low-resolution images \cite{kim2013vision}. While the approach is effective up to 5 meters, it could only recognize arm gestures with no capability to observe finger details. A different approach used a CNN model to capture spatiotemporal features for hand gestures over a sequence of images \cite{AlHammadi2020}. While providing high recognition rates, all approaches are limited to short- or long-range and cannot be applied to URGR.

Some approaches are based on a human pose estimation model often termed a skeleton model \cite{lai20163d,fu2019research,Gao2022,An2022}. Existing packages such as OpenPose \cite{Cao2019} and MediaPipe \cite{mediapipe2019} estimate landmarks of the human body including fingers. Some packages may use only an RGB camera \cite{Qiao2017} while others require also depth information \cite{Mazhar2018}. From estimated landmarks, one can infer about the exhibited gesture. Using a skeleton model provides easy access to hand pose and gesturing shapes. However, due to the limitations of these skeleton models, gesture recognition is only available in short- and long-range. While regular Convolutional Neural Networks (CNN) struggle to capture the structure of the human body in a frame, GCNs are prominent components in common skeleton models since they can learn the spatial relationship between body joints \cite{Aiman2024}. Hence, they are used observe mutual joint configurations to understand the overall gesture \cite{Li2019gcn,Fang2020}. Therefore, we integrate GCN in the proposed URGR approach.

Wearable technology is an alternative approach where some measurement device is attached to the user. With a wearable device, distance is only limited by communication with the receiving agent \cite{Tchantchane2023}. The notable method is the use of inertial sensors on a smartwatch to track motions of the arm and hand \cite{Wang2023}. However, inertial sensors alone can have difficulties to implicitly sense the state of the hand and, thus, are usually integrated with other sensors \cite{Alemayoh2022,Bongiovanni2023}. While inertial sensors directly observe mechanical motions of the arm, Electro-Myography (EMG) measures biometrics of indirect muscle activity \cite{Benalcazar2017}. EMG is another wearable approach where electrical currents in muscle cells are measured and provide some state of the hand \cite{Moin2020}. In \cite{Lian2017}, a wearable band on the forearm measured a 3-channel EMG signal during gestures. A model combining kNN and Decision Tree algorithms was trained to classify the signals into gestures. Similarly, EMG signals were used to train a convolutional recurrent neural network to control a robotic arm through gestures \cite{Kim2023}. In a different biometric approach, Force-Myography was used to sense muscle perturbations on the arm using simple force sensors and to classify signals to corresponding gestures \cite{Fora2021}. Other approaches include optical sensors for measuring blood volume \cite{Negreiros2022} and acoustic sensors on the wrist which measure tendon movements \cite{Siddiqui2017}. Another wearable approach is egocentric vision where a wearable camera is mounted on the head or chest of the user \cite{Bandini2023}. Hence, the camera has a first person perspective and observes the front of the user including the hands. In such a case, hand gestures can easily be observed from a very close range with almost no occlusions \cite{Alam2022}. While can provide a high recognition rate, the above wearable approaches may require designated and expensive hardware. Hence, they do not allow occasional users to interact with a robot. Also, biometric approaches may not provide generalized model transfer to various new users and additional data collection would be required.

To summarize, current gesture recognition approaches are either based on on-body devices or very limited in range. On-body devices limit occasional interaction with a robot and usually do not generalize well. On the other hand, using a camera on the robot enables gesture-based directives by occasional users and can generalize well to various users. However, existing methods for gesture recognition are not feasible in distances farther than 7 meters and usually limited indoors. 
These limitations of existing gesture recognition methods are mostly a combination of factors such as low resolution and lighting. When the distance between the camera and the subject increases, the resolution of the captured images decreases.
Indoor environments often present challenges related to lighting conditions. In poor lighting, shadows and increased distance from the camera, gesture recognition approaches encounter challenges of reduced image quality, higher noise levels and limitations in sensor capabilities, impacting their ability to distinguish between relevant gestures and background information. Other sources of reduced performance are interference from other users, cluttered background, limited field of view and dynamic environments.

\subsection{Super Resolution}
\label{sec:SR}

As discussed in the previous section, SR aims to enhance low-resolution images and improve distorted details. Several works have used image enhancement in gesture recognition \cite{Sundaramoorthy2021}. However, in the majority of the approaches a set of standard filters was applied to remove noise and improve the contrast of a gesture \cite{Li2020,Kaur2021}. However, the enhancements are implemented in close ranges where their need is not clear. Various other methods have been proposed for different applications such as satellite and aerial imagery, object detection in complex scenarios, medical imaging, astronomical images, forensics, face detection and recognition in security and surveillance, number plates reading and text analysis \cite{Nasrollahi2014,Anwar2020}. One approach is the Robust U-Net (RUNet) \cite{Hu2019} which is a variant of the well-known U-Net for segmentation \cite{ronneberger2015}. U-Net was initially used for image segmentation for biomedical applications where the image is contracted and expanded in a U-shaped path of convolutional layers. However, it was shown to be efficient for image quality improvement \cite{Lu2019}. A more advanced approach is the Enhanced Super-Resolution Generative Adversarial Networks (ESRGAN) \cite{Wang2019}. ESRGAN uses an adversarial process to generate realistic textures in a low-resolution image. Blind Super-Resolution Generative Adversarial Networks (BSRGAN) is an extended version of ESRGAN using different data and loss functions \cite{zhang2021}. Similarly, Real-ESRGAN uses a high-order degradation process to better simulate complex image corruptions \cite{Wang2021}. 

While efficient, it has been claimed that SR models are generally trained for specific tasks and may not perform well for different ones \cite{Anwar2020}. In particular, and to summarize, no SR model has been designated to the task of gesture recognition. In ultra-range, in particular, the prominent challenges are resolution drop, motion blur and noise which introduce complexities that existing SR models often struggle to address, leading to misinterpretations and decreased performance \cite{WangZhihao2019,Ye2023}. An SR model for URGR should be able to enhance image details including clear edge detection and higher fidelity in fine features, which are crucial for accurately interpreting gestures in low-resolution.

\section{Methods}
\label{sec:method}
\subsection{The Ultra-Range Recognition Problem}
\label{sec:problem_def}
The quality of images degrades with the increase of camera-to-subject distance $d$. The degradation stems from a combination of factors including the dimensions of the image sensor $S$, the optical characteristics of the camera lens $L$, and the distance $d$. Upon capturing an image, the quantity of light incident on the image sensor diminishes proportionally with the increase in $d$. This light attenuation results in a reduction of the signal-to-noise ratio (SNR) and a diminution of fine details within the image. Moreover, the sensor dimension $S$ significantly influences its light-capturing capacity where smaller sensors tend to yield images of inferior quality. 

Typically, when the object of interest is distant from the camera, zooming becomes necessary to perceive finer details at the expense of image sharpness and clarity. This results in reduced quality. Moreover, pronounced movements often present in long-distance interactions lead to motion blur in captured images. The presence of motion blur presents a significant obstacle to accurately extract 
information in fine detail. To cope with these challenges, the development of robust algorithms becomes imperative, capable of effectively mitigating motion blur and capturing meaningful patterns.

In this work, we focus on recognizing human gestures in the context of directing robots by, for instance, pointing and beckoning. Gesture recognition is required to be performed in a camera-to-gesture distance of up to 25 meters. In such a scenario, recognition of a blurry and difficult to distinguish hand in an image with many details is required. The image quality and large variance of gesture appearance at longer distances pose a complex challenge to the efficacy and precision of a Deep Neural Network (DNN) model. The model has limited features in the images that can facilitate the learning process of the DNN. Furthermore, factors such as diminished visual acuity, occlusions and perspective distortion contribute to ambiguity and diminished distinguishability of gestures. 

Formally, the problem addressed in this paper is as follows. Given an RGB image $\ve{J}_i$ where a single user is observed within it and with $d\leq25$ meters. The user can exhibit one out of $m$ gesture classes $\{\mathcal{O}_1,\ldots,\mathcal{O}_m\}$. Class $\mathcal{O}_1$ is the null class where no gesture is performed and the user conducts any other task. A gesture can be performed with either the right or left arm. We require a trained classifier that will recognize the gesture exhibited by the user in the image. In other words, a trained classifier will solve the following maximization problem
\begin{equation}
    \label{eq:P(O)}
    j^*=arg\max_j P(\mathcal{O}_j|\ve{J}_i),~\forall j=1,\ldots,m
\end{equation}
where $P(\mathcal{O}_j|\ve{J}_i)$ is the conditional probability for image $\ve{J}_i$ to be in class $\mathcal{O}_j$. Problem \eqref{eq:P(O)} is solved such that class $\mathcal{O}_{j^*}$ corresponds to the true gesture exhibited in image $\ve{J}_i$. It is important to emphasize that the URGR problem in our context uses only a single RGB image for recognition while we leave inference over a set of temporal images for future use.





\subsection{Data Collection}
\label{sec:data_collection}

A dataset $\mathcal{H}$ is collected for training the recognition model using a simple RGB camera. A set of $N$ images is collected by taking images of various users at different distances from the camera and in different environments. Environments can include indoor and outdoor ones. The users exhibit any of the pre-defined gestures of classes $\{\mathcal{O}_1,\ldots,\mathcal{O}_m\}$. To maintain a nearly uniform distribution, an equal amount of images is taken for each gesture class. In addition, a uniform distribution of samples is taken along the distance from the camera. For such, a long measuring tape was used such that an equal amount of samples is taken for each distance in intervals of one meter and in the range $d\in[0,25]$ meters. Each image $\ve{J}_i$ is labeled with the class index $o_i\in\{1,\ldots,m\}$ and distance $d_i\in[0,25]$. Consequently, the resulted training dataset is of the form $\mathcal{H}=\{(\ve{J}_i,o_i, d_i)\}_{i=1}^N$. 



\subsection{Image Quality Improvement}
\label{sec:hq}

In this part, we propose a novel model termed \textit{\hq} for solving the Single Image Super-Resolution (SISR) problem where a trained model infers a high-resolution image from a low-resolution one.

\subsubsection{Pre-Processing}
\label{sec:preproc}

A subset of dataset $\mathcal{H}$ is used to train the \hq model. Since \hq is aimed to improve the quality of zoomed-in images, we require images with sufficient quality for ground-truth labeling. Hence, we define subset $\mathcal{S}\subset\mathcal{H}$ where only samples taken in the range $d\in[2, 8]$ meters are used. The degradation becomes predominant when $d>8~m$ and it is not possible to produce image labels with sufficient quality for training.  

\begin{figure}
    \centering
    \begin{tabular}{cc}
       \includegraphics[width=0.35\linewidth]{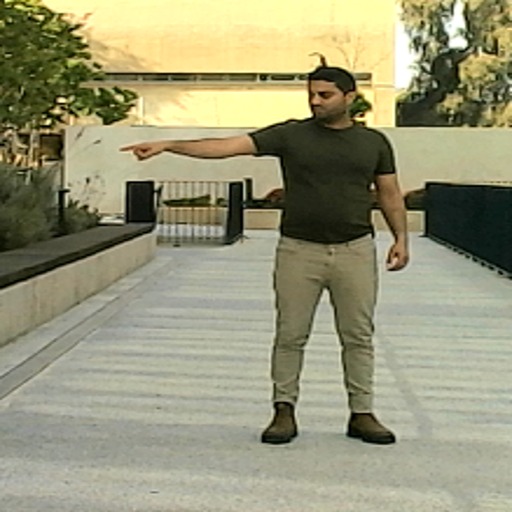}  & \includegraphics[width=0.35\linewidth]{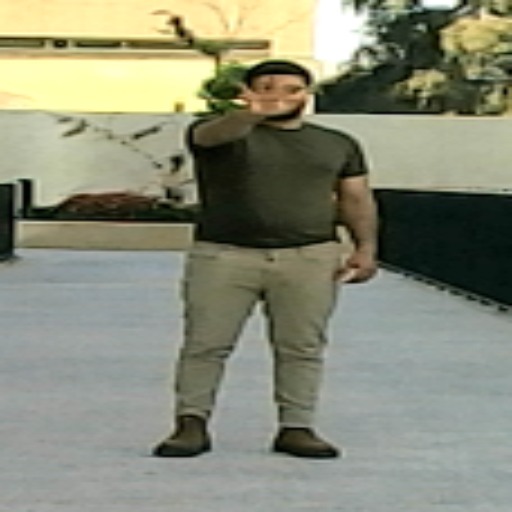} \\
       (a) & (b) 
    \end{tabular}
    \caption{Image examples of (a) pointing and (b) stop gestures showing different widths of the user. Hence, pixels around the user are added in order to maintain a constant image proportion.}
    \label{fig:proportions}
\end{figure}

In order to emulate the appearance of ultra-range images after focusing on the user, each image $\ve{J}_i\in\mathcal{S}$ is degraded by the following manipulations. First, we require images focused on the user (i.e., the user is centered in the frame and covers most of the image) while the gesture is visible. Using a \textit{You Only Look Once} version 3 (YOLOv3) \cite{redmon2018yolov3} bounding box, the user is detected within the image yielding a bounding box of size $w\times h$. However, the bounding box may not always bound the entire human body completely and can miss vital parts such as the gesturing arm. Hence, an extension of the bounding box is required while maintaining a constant image proportion. However, the width of the user may be different between various gestures. For instance, as seen in Figure \ref{fig:proportions}, the width of the user when pointing is much larger when exerting a stop gesture. Hence, the pixel thickness that will be added around the bounding box is $\frac{b}{a}$ where $b$ is the diagonal length of the bounding box and parameter $a$ is a pre-defined user-to-image ratio. Then, the image is cropped around the extended bounding box yielding an image size of $(w+\frac{b}{a})\times(h+\frac{b}{a})$. Finally, the cropped product is resized to resolution $512\times512$ so to have a unified size in the dataset. This operation will results in the user appearing approximately at the same size within the image irrespective of the distance from the camera, while maintaining a constant size and ratio for the model.

After focusing on the user in the image and cropping, the image $\ve{J}_{i}$ is deliberately degraded in order to emulate an image taken from a long distance. We have developed a degradation process comprising a series of transformations and the application of multiple filters. These filters collectively contribute to the creation of a nuanced shading effect around the object of interest and detail corruption, thereby emulating the characteristics of long-distance perspective. The formalized expression for acquiring a degraded image $\tve{J}_{i}$ is given by 
\begin{equation}
\tve{J}_{i} = F_{compress} (F_{sharpen}(F_{smooth}(\ve{J}_{i}))).
\end{equation}
Function $F_{smooth}$ is a \textit{smoothing} operation where a spatial filtering technique reduces the high-frequency noise. In practice, a Gaussian filter with a kernel size of $5\times5$ is applied. Function $F_{sharpen}$ is a conventional \textit{sharpening} operator with a $3\times3$ kernel, designed to enhance image sharpness, particularly in high-frequency regions. Finally, the images undergo \textit{compression} with function $F_{compress}$ by encoding it using the JPEG format. This leads to a reduction in fine details and fidelity.
Through experimentation, we have found that the combination of sharpening and compression operations, while degrading the quality of details, also creates the shadowing effect in the high-contrast regions that surround the user.
In conclusion, each image $\ve{J}_i\in\mathcal{H}$ in the range $d\in[2, 8]$ meters is passed through the above process to generate dataset $\mathcal{S}=\{(\tve{J}_i,\ve{J}_i)\}_{i=1}^M$ with $M$ samples.

\begin{figure*}
    \centering
    \includegraphics[width=\textwidth]{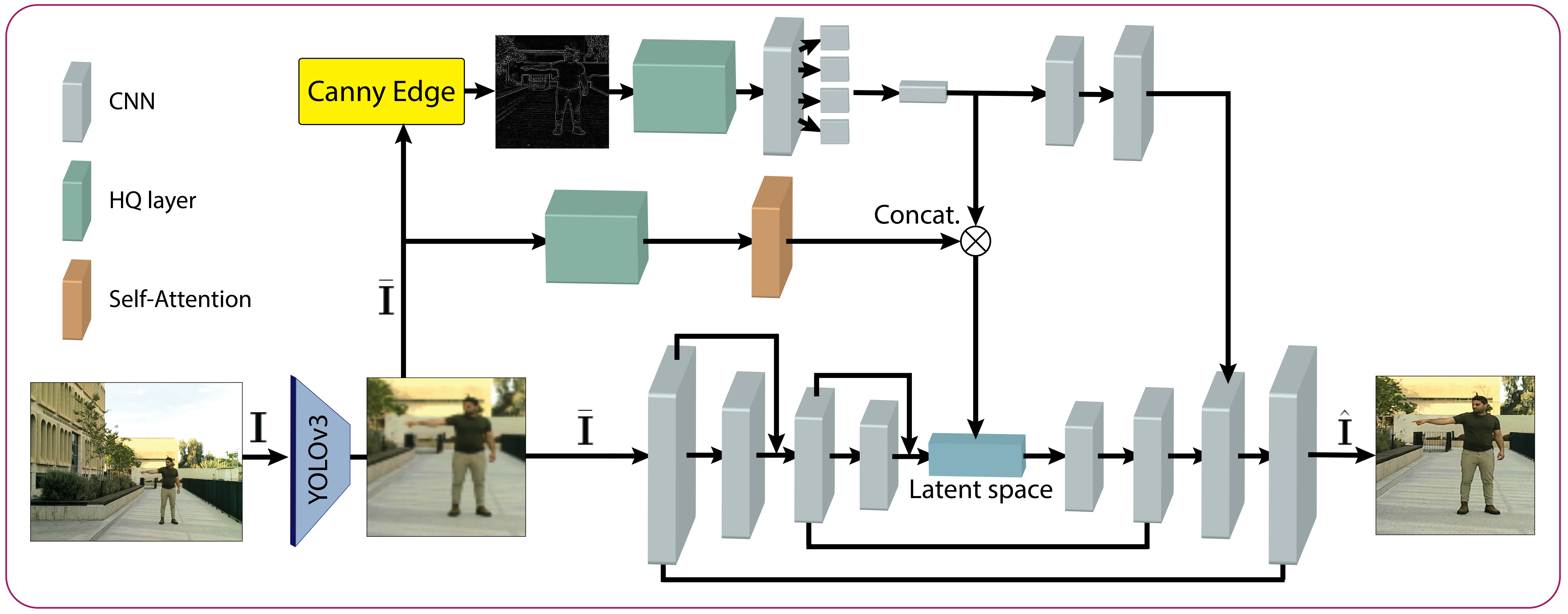} 
    \caption{Illustration of the \hq model focusing on the user. A cropped image is the input to three pathways yielding a quality improved image $\hve{I}$.}
    \label{fig:SR_gesture}
\end{figure*}
\begin{figure}
    \centering
    \includegraphics[width=0.55\linewidth]{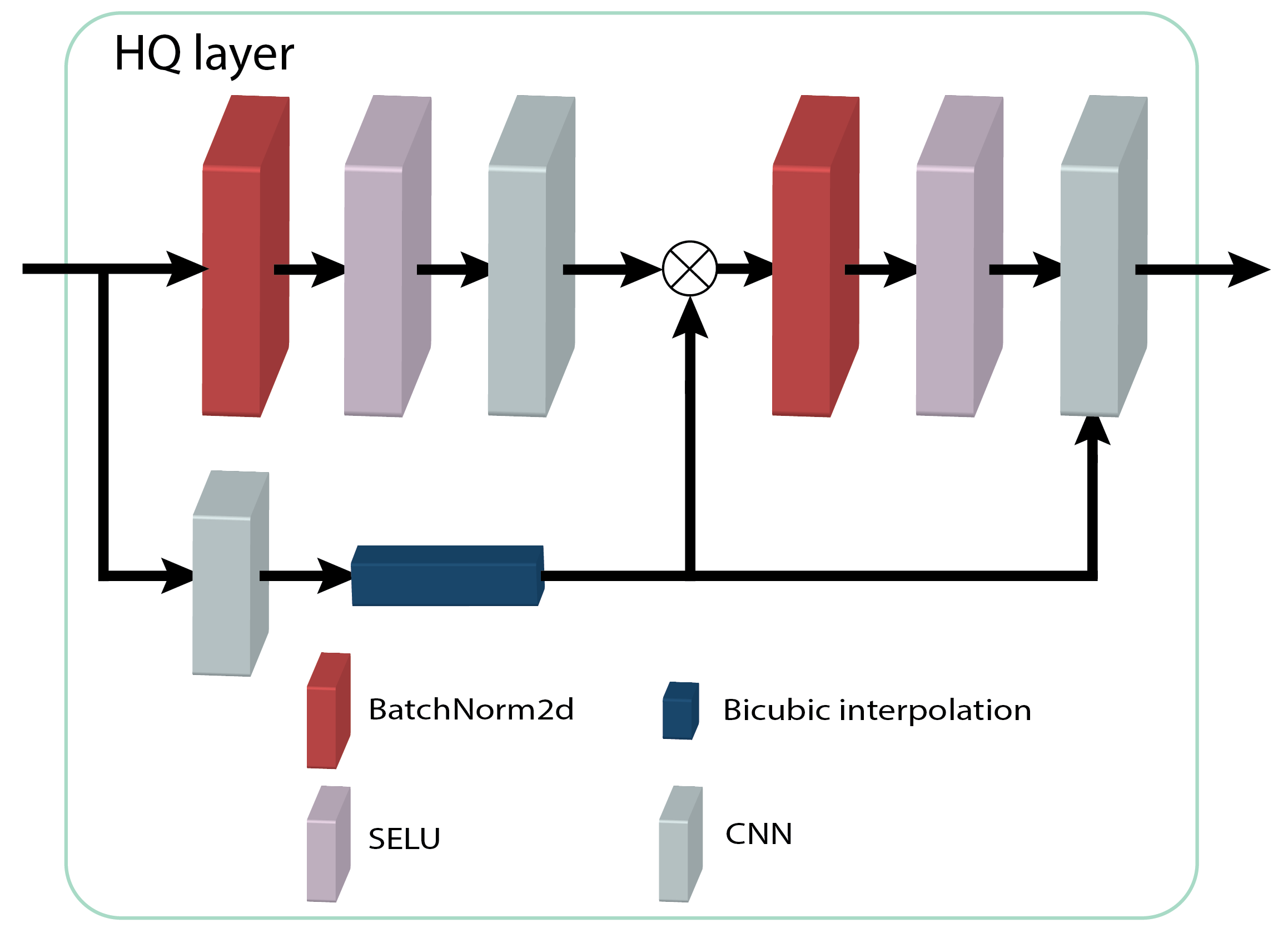} 
    \caption{Architecture of the HQ layer used in the \hq (Figure \ref{fig:SR_gesture}) for improving image quality of a user in ultra-range.}
    \label{fig:SR_HQ}
\end{figure}
\subsubsection{Super-Resolution Model}
\label{sec:SR}

Upon capturing an image $\ve{I}$ in ultra-range, the \hq model will optimize image quality in the region of the visible user. The \hq is illustrated in Figure \ref{fig:SR_gesture}. The architecture of the HQ-Net model is guided by a fusion of rigorous design choices and empirical scrutiny. A main component of HQ-Net is the HQ-layer seen in Figure \ref{fig:SR_HQ} and is composed of a series of convolutional layers, batch normalization and Scaled Exponential Linear Unit (SELU) activation functions. The HQ-layer is structured by bifurcating the input. The first branch proceeds through designated convolutional layers, while the second branch undergoes a single convolutional layer followed by Bicubic interpolation. The outputs of both branches are element-wise multiplied and fused with the final convolutional layer within the network. 

In the first step of HQ-Net, the user is localized within the image using YOLOv3 and cropped-out to a sub-image $\bve{I}$ simulating its proximity to the camera. With $\bve{I}$, a three-path processing scheme is employed. The choice of hyper-parameters and activation functions was conducted upon optimization. In the initial pathway, the Canny edge detection algorithm is used to discern prominent edges within $\bve{I}$. This enables HQ-Net to better identify important structural features in the image during the enhancement process. By detecting edges, HQ-Net can focus on preserving these critical features while reducing noise and irrelevant details. The Canny edge detector parameters were adjusted empirically with a Gaussian filter standard deviation of 1.4 along with low and high thresholds of 0.1 and 0.3, respectively. The edge image then passes through the HQ layer and a set of convolutional layers yielding a latent vector of size 284. 

In the second pathway, cropped image $\bve{I}$ is passed through an HQ layer followed by a self-attention mechanism that outputs a latent vector of size 552. The self-attention plays a significant role in understanding the spatial relationships and context within the image. The integration of self-attention mechanisms empowers the network to prioritize pertinent details by assigning elevated weights to significant regions of the image. Hence, it enables the model to allocate varying degrees of importance to distinct segments of the input image. By computing attention scores within the input, the model discerns which features are most pertinent for the given task, prioritizing salient regions like object boundaries or distinctive patterns while attenuating less relevant areas. Moreover, this mechanism facilitates noise reduction by capturing extensive dependencies and correlations spanning different portions of the image. Through dynamic adjustment of attention weights, the model accentuates semantically meaningful aspects while disregarding extraneous background clutter, thereby refining its focus on the primary task. Overall, the self-attention prioritizes relevant details, reduces noise and eliminates distractions, collectively leading to a significant improvement in the overall quality of the processed image. The third pathway has an auto-encoder framework with several convolutional layers and skip connections. The inclusion of skip connections facilitates the flow of information between encoder and decoder layers, enabling the network to effectively capture both low-level and high-level features. The layers map image $\bve{I}$ to a latent space of size 2048. Outputs from the first two pathways are concatenated into the latent space with a total size of 2,884 (2,048 + 284 + 552) for further expansion with an additional set of convolutional layers and skip-connections. The output is the quality improved image $\hve{I}$.

This novel network architecture of \hq is tailored to scenarios where image fidelity is compromised by distant capture and subsequent crop operations. \hq is trained using dataset $\mathcal{S}$ such that the input is the degraded image $\tve{I}_i$ with corresponding output label $\ve{I}_i$. Training involves the minimization of a pixel-to-pixel Mean square Error (MSE) loss function to compare between the improved image $\hve{I}_i$ and the ground truth one $\ve{I}_i$. MSE measures the average of the squares of each pixel error and is given by
\begin{equation}
    \label{eq:MSE}
    \text{MSE}=\frac{1}{n_xn_y} \sum_{j=1}^{n_x} \sum_{k=1}^{n_y} ( \ve{i}_{i,j,k}-\hve{i}_{i,j,k} )^2
\end{equation}
where $n_x$ and $n_y$ are the number of rows and columns in the images, respectively, and $\ve{i}_{i,j,k}$ is the $(j,k)$ component of image $\ve{I}_i$. In such a way, MSE evaluates pixel dissimilarities between the output and ground truth images.

\subsection{URGR Model }
\label{sec:Long}

Gesture recognition encompasses a set of methodologies and algorithms designed to identify and interpret human gestures depicted within images. As discussed in Sections \ref{sec:introduction}-\ref{sec:relatedwork}, state-of-the-art approaches for gesture recognition are designed to identify human gestures in close proximity of up to 7 meters. While image enhancing using \hq is beneficial, it does not provide a complete solution for observing fine details. Hence, given a quality improved image $\hve{I}$ originally taken from a distance of up to 25 meters, our proposed Graph-Vision Transformer (\gv\!\!) will output a solution to \eqref{eq:P(O)}. 

As previously discussed, the greater the distance between the user and the camera, the more challenging it becomes for the model to accurately discern the executed gesture. This challenge is primarily attributed to the substantial presence of irrelevant background objects in such scenarios. In order to mitigate the influence of background objects, user crop-out and quality improvement are exerted. Hence, for training, dataset $\mathcal{H}$ is pre-processed such that each image $\ve{I}_i\in\mathcal{H}$ is cropped (with YOLOv3) around the recognized user while maintaining the same image proportion as in Section \ref{sec:preproc}. Then, the cropped image is passed through the \hq for quality improvement yielding a modified training dataset $\hat{\mathcal{H}}=\{(\hve{I}_i,o_i, d_i)\}_{i=1}^N$ where $\hve{I}_i$ is the processed image.

The \gv model combines the power of Graph Convolutional Networks (GCN) \cite{Ullah2019} with the expressiveness of a modified Vision Transformers (ViT) \cite{dosovitskiy2020image}. Thus, it could effectively process and analyze structured data, such as images, with inherent graph-like relationships while leveraging the self-attention mechanism of ViT in order to capture fine-grained visual features. The GViT structure is illustrated in Figure \ref{fig:GVIT}. We briefly present the notion of the GCN and ViT models followed by a discussion on GViT.

\subsubsection{Graph Convolutional Networks} GCN has emerged as a versatile and powerful tool for image processing, offering distinct advantages over the traditional Convolutional Neural Network (CNN). GCN is applied on undirected graphs in order to capture intricate relationships among nodes in the graph. Let $\mathcal{G}=(V,E)$ be an undirected graph where $V$ and $E$ are the set of nodes and edges connecting them, respectively. In GCN, $\mathcal{G}$ goes through a set of Graph Convolutional (GC) layers. In a GC layer, each node aggregates information from its neighboring nodes while considering both its own features and the features of its neighbors. Let $\ve{h}_x^{(k)}$ by the feature representation of node $x\in V$ in layer $k$. The aggregation $\ve{h}_i^{(k+1)}$ of node $x$ in the next layer is acquired through a weighted sum
\begin{equation}
    \label{eq:GClayer}
    \ve{h}_x^{(k+1)} = \sigma \left( \sum_{y \in N(x)} W^{(k)} \ve{h}_y^{(k)} \right) 
\end{equation}
where $N(x)\subset V$ is the subset of neighboring nodes to $x$, $\ve{W^{(k)}}$ is a learnable weight matrix of layer $k$, and $\sigma$ is some activation function.

In the field of image analysis, a graph can represent relationships between pixels or image regions \cite{Li2019,Vasudevan2023}. Each pixel can be a node and edges can represent the spatial adjacency of pixels. Then, the GCN can utilize such a structure to model intricate relationships among pixels and capture spatial dependencies within the image. It has the ability to capture long-range dependencies and relationships between non-adjacent pixels. Furthermore, GCN has fewer parameters compared to a traditional CNN, making it more memory-efficient and easier to train.

\subsubsection{Vision Transformer} Transformers have been a ground-breaking architecture for natural language processing \cite{Vaswani2017}. In general, Transformers enable the learning of long-range dependencies in sequential data. While an image is not considered sequential data, ViT is an extension of the Transformer for learning dependencies and intricate spatial relationships between regions in an image. Hence, the key steps in ViT begin with patch extraction, where input images are divided into non-overlapping patches, each treated as a sequence of feature vectors. To introduce spatial information, positional encoding is added to the patches. 

Linear embeddings of the patches are considered as tokens and fed into the Transformer encoder of the ViT. The encoder is composed of a stack of multi-head self-attention layers. These layers capture long-range dependencies between the patches. The self-attention mechanism evaluates the hierarchical importance of each patch with the assistance of Softmax functions. 
The outputs of the Transformer encoder are passed to a classification head implemented by a  global average pooling and feed-forward neural network. This hierarchical approach is considered highly beneficial for complex visual recognition. 


\begin{figure*}
    \centering
    \begin{tabular}{cc}
        \includegraphics[width=\textwidth]{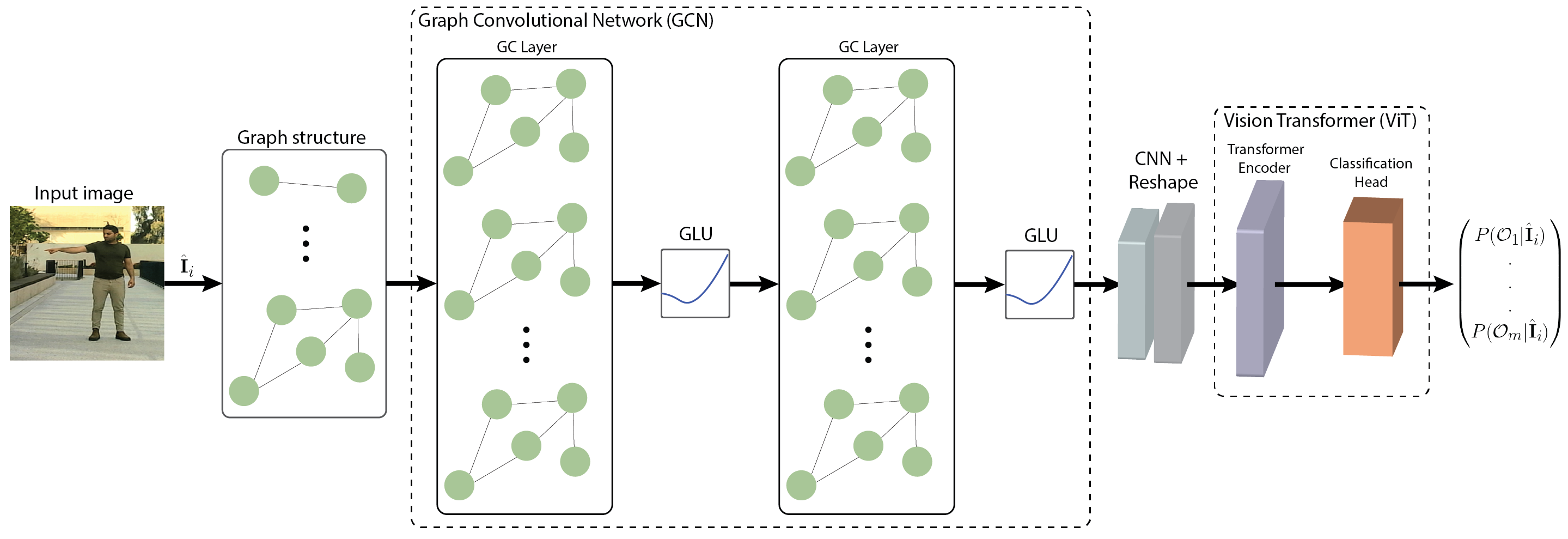} 
    \end{tabular}
    \caption{The proposed \gv model for URGR. The model reconstructs the image into a graph structure followed by GCN and ViT models.}
    \label{fig:GVIT}
\end{figure*}

\subsubsection{Graph-Vision Transformer (GViT)} \gv combines the benefits of GCN and ViT such that it can effectively model both local and global dependencies within the structured data. We are motivated by the ability of GCN to capture fine-grained spatial relationships while ViT can effectively model global context and semantics. Hence, this hybrid approach could allow for a more comprehensive analysis of the image at multiple scales, potentially leading to improved performance compared to using either model individually. A processed image $\hve{I}_i\in\hat{\mathcal{H}}$ is passed through a sequence of two GC layers and a ViT as illustrated in Figure \ref{fig:GVIT}. As discussed above, the input to the standard ViT is a linear embedding of the patches. In \gv\!\!, however, we modify the ViT such that the GC layers are its embedding input. First, image $\hve{I}_i$ is converted into a graph structure by iterating over all pixels. Each pixel of the image corresponds to a node in the graph \cite{Bae2022}. Hence, the total number of nodes in the initial graph is equal to the total number of pixels in the image. For each pixel, edges are established with its neighboring pixels including diagonal ones. These edges construct set $E$ and define the connectivity pattern of graph $\mathcal{G}$. In practice, edge set $E$ can be mathematically described through the concept of an adjacency matrix. Each element $E_{i,j}$ of this matrix is given by
\begin{equation}
    E_{i,j} = \begin{cases}
    1, & \text{if } |x_i - x_j| \leq 1 \text{ and } |y_i - y_j| \leq 1 \text{ and } \ve{a} \neq \ve{b} \\
    0, & \text{otherwise}
    \end{cases}
\end{equation}
and represents the connection strength between nodes $i=(x_i,y_i)$ and $j=(x_j,y_j)$, corresponding to individual pixels in the image. This formulation describes the construction of the image graph and its corresponding adjacency matrix.



Using the initial graph structure, GCN propagates information with graph convolutions in order to capture dependencies in the image. We reformulate \eqref{eq:GClayer} and write it in a matrix form as
\begin{equation}
    \label{eq:GClayer_re}
    H^{(k+1)} = \sigma(D^{-\frac{1}{2}} E^{(k)} D^{-\frac{1}{2}} H^{(k)} W)
\end{equation}
where $H^{(k)}$ is the input to layer $k$ such that $H^{(1)}=\hve{I}$, $E^{(k)}$ is the adjacency matrix of layer $k$ and $D$ is the degree matrix containing the sum of weights of the nodes in a graph. The graph is propagated through two GC layers defined in \eqref{eq:GClayer_re} with activation function $\sigma()$ implemented by the Gated Linear Unit (GLU). GLU is a gating mechanism following a convolutional operation to control the flow of information \cite{Dauphin2017}. The key advantages of the GLU activation function in image recognition tasks are capturing complex patterns, effective feature learning and the reduction of overfitting. In addition to the GLU, dropout is added between the GC layers to make the model more general and to mitigate overfitting. Within a GCN layer, information is aggregated from neighboring nodes yielding updates to the features of each node based on this aggregated information. This process involves propagating information through the graph edges according to the learned weights of the GCN layer. Therefore, while the initial graph structure captures pixel-level connectivity, the operations within the GCN layer modify the node features based on the learned graph convolutional operations, resulting in updated representations that are suitable for subsequent processing tasks such as gesture recognition.


Instead of the flatten projection operator in the original ViT, the output nodes of the GCN are passed through a single convolutional layer to reduce the number of image channels and adapt to the optimized input size of the ViT. Finally, the ViT component takes the output of GCN effectively replacing the linear embedding part of the traditional ViT. The ViT model extracts meaningful features from the graph representation using the self-attention mechanism. Following the self-attention step, the output is further processed through the classification head to acquire a probability distribution $(P(\mathcal{O}_1|\hve{I}_i),\ldots,P(\mathcal{O}_m|\hve{I}_i))$ over the objects. Then, the recognized gesture is the solution to \eqref{eq:P(O)}. \gv is trained using the cross-entropy loss function. In the next section, we demonstrate the effectiveness of the proposed \gv through experiments, showcasing its ability to achieve state-of-the-art results in recognizing human gestures in both short and long ranges of up to 25 meters.

\section{Model Evaluation}
\label{sec:Evaluation}
\begin{figure}
    \centering
    \includegraphics[width=\linewidth]{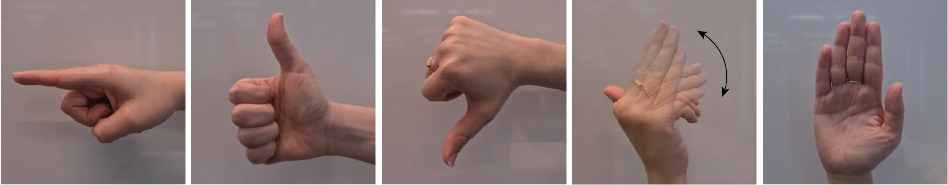}
    \caption{Five directive gesture classes considered in this paper, from left to right: pointing, thumbs-up, thumbs-down, beckoning and stop. A sixth class is the null one where the user does not exhibit any gesture.}
    \label{fig:gestures}
\end{figure}

In this section, we present the testing and analysis of the proposed URGR framework. Without loss of generality, we chose to focus on six directive gesture classes seen in Figure \ref{fig:gestures} including: pointing, thumbs-up, thumbs-down, beckoning and stop. A sixth class is the null one where the user does not exhibit any gesture and can perform any other task. As discussed previously, we analyze the ability of the proposed model, along with other known methods, to identify the gestures in a range of up to 25 meters. For a basis of comparison, we first present results of recognition by human participants directly observing the gestures from a long distance. Then, a comparative evaluation of the proposed model is discussed. All experiments were carried out on a Linux Ubuntu (18.04 LTS) machine with an Intel Xeon Gold 6230R CPUs (20 cores running at 2.1GHz) and four NVIDIA GeForce RTX 2080TI GPUs (each with 11GB of RAM). 

In this section, we first provide reference results of gesture recognition by humans. Then, we present the datasets collected for training the models followed by comparative evaluation of HQ-Net and GViT. We then analyze the data requirements of the models and evaluate specific edge case scenarios for recognition. 

\subsection{Human Gesture Recognition}

Irrelevant to the hardware and any learning method, we wish to examine the ability of humans to recognize gestures in ultra-range. Hence, an experiment was designed and conducted in which a human demonstrator presented the gestures from various distances. Also, the six gestures were presented in a uniform distribution and in multiple indoor and outdoor environments. Ten participants of different age ranges were asked to name the gesture they recognize made by the demonstrator. Participants requiring distance glasses were asked to wear them. The distance range of $[2,25]$ meters was partitioned into five ranges and, for each range and participant, ten recognition trials were made. Distance ranges were presented in an ascending order from short to ultra range. In addition, gestures were presented to the user in various arm positions. Table \ref{tb:human} summarizes the human recognition results. The results show, as expected, that short and long-range recognition is much more successful than ultra-range ones across all participants. In particular, the probability for a human to recognize a gesture in the range of $[19,25]$ meters is approximately $0.5$.

\begin{table*}[]
\centering
\caption{Human recognition of gestures in various distance ranges}
\label{tb:human}
\begin{tabular}{cccccccc}
\toprule
Participant & \multirow{2}{*}{Age} & \multicolumn{5}{c}{Success rate (\%) in distance range} & Total \\ \cmidrule{3-7}
\# &  & 2-7 (m) & 8-12 (m) & 13-18 (m) & 19-22 (m) & 23-25 (m) & (\%) \\ 
\midrule
1   & 20-25       & 100   & 100    & 100   & 70      & 70      & 88    \\
2   & 20-25       & 100   & 100    & 80    & 70      & 60      & 82    \\ 
3   & 25-30       & 100   & 100    & 80    & 70      & 40      & 78    \\
4   & 25-30       & 100   & 100    & 100   & 70      & 60      & 86    \\
5   & 35-40       & 100   & 100    & 90    & 60      & 50      & 80    \\
6   & 35-40       & 100   & 90     & 70    & 60      & 40      & 72    \\
7   & 50-60       & 100   & 100    & 80    & 50      & 40      & 74    \\
8   & 60-70       & 100   & 100    & 80    & 60      & 40      & 76    \\
9   & 60-70       & 100   & 100    & 90    & 60      & 60      & 82    \\
10  & +70         & 100   & 90     & 70    & 50      & 20      & 66    \\
\midrule
All &             & 100 & 98 & 84 & 62 & 48 & 78.4 \\
\bottomrule
\end{tabular}
\end{table*}

\subsection{Datasets}

As discussed in Section \ref{sec:data_collection}, dataset $\mathcal{H}$ is collected for training both \hq and \gv models. The dataset was collected using a simple web camera yielding images of size $480\times640$. In addition, 16 participants contributed data in various indoor and outdoor environments, and in a uniform distribution of data. All participants used only their right arm to exhibit gestures. Nevertheless, we will further analyze the generalization to the recognition of left arm gestures. 
The collection yielded 58,000 samples per gesture class and a total of $N=347,483$ labeled samples in $\mathcal{H}$. Another set of 10,109 labeled images was taken as a test set in environments and with users not included in the training. As presented in Section \ref{sec:preproc}, a subset $\mathcal{S}\subset\mathcal{H}$ is taken and used for training the \hq model. The subset includes samples in the range of $[2,8]$ meters and is of size $M=191,574$. Here also, $6,319$ independent images were taken as a test set for evaluating the \hq model. After the training of \hq using $\mathcal{S}$, all $N$ samples in $\mathcal{H}$ were pre-processed to $\hat{\mathcal{H}}$ by focusing on the user and improving quality with \hq\!.

\subsection{Existing gesture recognition models}
\label{sec:existing_models}

We evaluate several known gesture recognition models that are publicly available. These include open-source packages such as SAM-SLR \cite{Jiang2021}, MediaPipe Gesture Recognizer \cite{mediapipe2019} and OpenHands by OpenPose \cite{Cao2019}. In these models, gesture classes that align with our model are thumbs-up, thumbs-down and stop. Hence, we evaluate the recognition success rate with only these gesture classes over our test set in ultra-range distance. Table \ref{tb:gestures_framework} presents the success rate of these models for ranges 0-5 meters and 5-25 meters, and for the entire range. Since the three models are specifically designed for the former range, high recognition rate is achieved. However, farther distances are hardly recognizable with a poor success rate. Particularly, the majority of successful recognitions in the 5-25 meters range are in the closer region while completely failing with distance larger than 7 meters. Designed for high detail at close distances, these models are not suited for the low-resolution, blurry images with cluttered backgrounds encountered in ultra-range recognition tasks.



\begin{table}
\centering
\caption{Gesture recognition of several existing models with respect to distance range}
\label{tb:gestures_framework}
\begin{tabular}{lcccc}\toprule
        \multirow{2}{*}{Models} & \multicolumn{3}{c}{Success rate (\%)}  \\
        & 0-5 m & 5-25 m & Total  \\
        \midrule
        SAM-SLR                         & 88.4 & 21.4 & 34.6 \\
        MediaPipe Gesture Recognizer    & 90.8 & 23.5 & 38.2 \\
        OpenHands                       & 92.4 & 26.1 & 42.7 \\
\bottomrule
\end{tabular}
\end{table}

\subsection{Evaluation of Image Quality Improvement}
\label{sec:IQI_eval}

As discussed in Section \ref{sec:hq}, model \hq is used to improve the quality of the observed user in the image prior to the recognition of a gesture. The model is trained with dataset $\mathcal{S}$ where a degraded image $\tve{J}_i$ is mapped to its original qualitative image $\ve{J}_i$. All hyper-parameters of HQ-Net including the architecture described in Section \ref{sec:hq} were optimized using Ray-Tune \cite{liaw2018tune} which efficiently searches through a large hyper-parameter space. Ray-Tune facilitates parallel and distributed training, thereby expediting the experimentation process through the concurrent utilization of multiple computational resources. This capability proves particularly advantageous when dealing with large-scale models like HQ-Net. In addition to the optimal architecture, the optimal training parameters included learning rate of 0.00485, batch size of 16, weight decay 0.0787 and dropout rate of 0.4.

We compare \hq to various state-of-the-art models including Autoencoder \cite{Zeng2017}, U-Net \cite{ronneberger2015}, U-Net++ \cite{zhou2019unet++}, Robust U-Net \cite{Hu2019}, ESRGAN \cite{Wang2019} and BSRGAN \cite{zhang2021}. The baseline is a simple Autoencoder where the image is encoded and reconstructed to the original variant. U-Net++ is an extension of the U-Net architecture with additional skip-connections to improve accuracy. The other benchmark models were discussed in Section \ref{sec:SR}. The hyper-parameters of these models were also optimized using Ray-Tune.


All compared models were trained using MSE loss function \eqref{eq:MSE}. In addition, we evaluate the trained models with the Peak Signal-to-Noise Ratio (PSNR). PNSR is a quality assessment metric for image restoration given by 
\begin{equation}
    \text{PNSR}=10\log_{10}\frac{L^2}{\text{MSE}}
\end{equation}
where $L$ is the maximum pixel value \cite{Chen2022}. Hence, PNSR is a measure of the maximum error in the image and is directly related to the MSE. The higher the PSNR value, the greater the reconstructed image resembles the ground truth one.

\begin{figure}
    \centering
     \includegraphics[height=0.9\textheight]{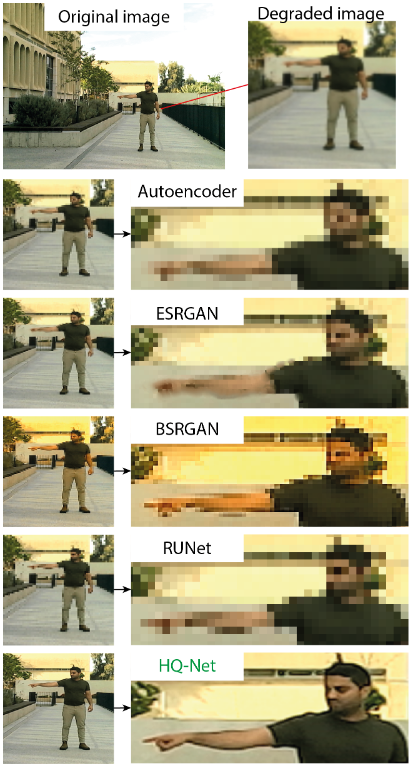} 
    \caption{Demonstration of a degraded image taken from a distance of 9 meters and improved using \hq and other methods. The original image is cropped using YOLOv3 and then degraded using the processing steps described in Section \ref{sec:preproc}.}
    \label{fig:SR9m}
\end{figure}

\begin{figure}
    \centering
     \includegraphics[height=0.9\textheight]{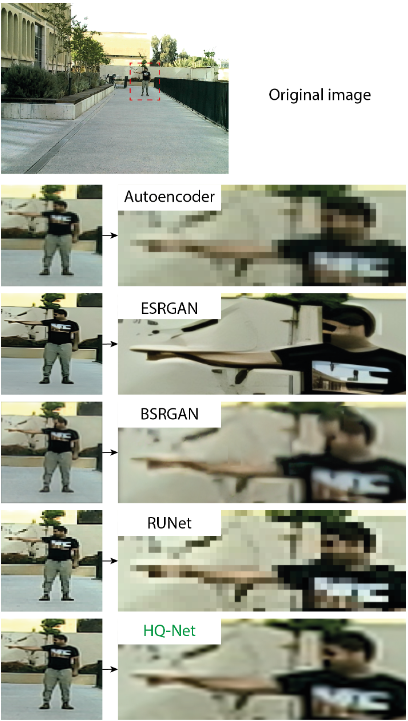} 
    \caption{Quality improvement example of an image taken from a distance of 25 meters. The user is identified using YOLOv3 and cropped-out. Then, it is improved using \hq and other methods.}
    \label{fig:SR25m}
\end{figure}

The comparison is done by improving the quality of the input image using each of the SR methods, and then classifying using GViT. The comparative results are presented in Table \ref{tb:Quality} by reporting both MSE losses and PSNR values for all methods over the test set. The results clearly show the dominance of \hq over the existing models. Figures \ref{fig:SR9m} shows an image captured from a distance of 9 meters and intentionally degraded as described in Section \ref{sec:preproc} for demonstration. Then, the degraded image is improved using \hq and the existing methods. Visually, \hq shows the best reconstruction of the image. On the other hand, Figure \ref{fig:SR25m} shows an example of an image taken from a distance of 25 meters and directly improved. \hq provides the best detail improvement and is further used for enhancing image quality before exerting a gesture recognition model. 

The architecture of HQ-Net incorporates several key features that contribute to its exceptional performance in image quality improvement. Compared to other methods, HQ-Net exhibits a more intricate architecture, allowing it to capture finer details and nuances within images. This heightened complexity enables HQ-Net to effectively reconstruct the image from a degraded one. Moreover, the integration of advanced techniques, such as the self-attention mechanism, provides the network with distinct advantages in performing image enhancement tasks. Additionally, the utilization of the Canny edge detection algorithm further enhances performance by providing accurate edge information. This enables HQ-Net to better identify and preserve important structural features in the image during the enhancement process. Furthermore, an examination of the architecture reveals that connecting the layers in the Autoencoder, built upon the methodology of skip connections, significantly enhances the model's capability for feature extraction. Of particular significance is our focus on the intricate inter-layer connections inherent within the Autoencoder architecture and illustrated in Figure \ref{fig:SR_HQ}.
This new version of connections between layers facilitates the flow of information between layers, allowing the model to bypass certain processing stages and retain important visual features. This design choice contributes to an ability to accurately reconstruct images while preserving crucial details. Nevertheless, some of the existing methods seem visually sufficient for gesture recognition. Hence, we further analyze their use for classification to justify the use of HQ-Net.

\begin{table}[]
\centering
\caption{Comparison of image quality improvement by various models}
\label{tb:Quality}
\begin{tabular}{lcccc}\toprule
        \multirow{2}{*}{Models}  &~~~~~~& \multicolumn{3}{c}{Model Score} \\\cmidrule{3-5}
        && MSE Loss &~~~~~~& PSNR (dB) \\\midrule
        Autoencoder  && 0.401 && 13.67  \\
        Unet  && 0.305 && 14.66 \\
        Unet++  && 0.289 && 15.01 \\
        RUNet && 0.198 && 21.77 \\
        ESRGAN  && 0.239 && 15.67 \\
        BSRGAN && 0.216 && 22.31 \\
        \rowcolor{lightgray}
        \hq && 0.019 && 34.45 \\
\bottomrule
\end{tabular}
\end{table}

\subsection{Gesture Recognition Model}

Once the image is cropped and improved using \hq, it would go through the \gv for classification of the gesture. We compare the proposed \gv model with various other models including a standard CNN, DenseNet \cite{huang2017densely}, EfficientNet \cite{tan2019}, GoogLeNet \cite{szegedy2015going}, Wide Residual Networks (WideResNet) \cite{Zagoruyko2016} and Visual Geometry Group (VGG) \cite{simonyan2014very}. Along with the proposed combination of GCN and ViT, performance is evaluated for each separately. The standard CNN was optimized to include five convolutional layers followed by max-pooling to down-sample the features, five fully-connected layers with ReLU activation functions and a Sigmoid function at the output. DenseNet and WideResNet are deep learning models that use densely connected blocks and residual connections, respectively, to improve accuracy and reduce overfitting. EfficientNet is a CNN model which uniformly scales the depth, width and resolution of the model to achieve a good balance between model complexity and performance. Similarly, GoogLeNet is also a CNN that has a unique inception module with multiple filters in parallel. VGG is a popular model for object recognition and segmentation which also uses a deep convolutional architecture. For the standard models, we compare results when fine-tuning (FT) the model or having a new train (NT) from the ground up, with the training data in $\mathcal{H}$.
\begin{table*}
\centering
\caption{Gesture recognition success rate with various methods with and without user crop-out and quality improvement using HQ-Net, and while fine-tuning (FT) existing models or having an entirely new train (NT)}
\label{tb:ClassificationWithoutProcessing}
\begin{tabular}{lcccccc}\toprule
    \multirow{2}{*}{Models~~~~~~~~~~~~~~~~~}  & \multicolumn{2}{c}{No pre-processing} & \multicolumn{2}{c}{w/ crop-out} & \multicolumn{2}{c}{w/ crop-out \& \hq} \\\cmidrule{2-3}\cmidrule{4-5}\cmidrule{6-7}
                    & FT (\%) & NT (\%) & FT (\%) & NT (\%) & FT (\%) & NT (\%) \\\midrule
    CNN             & -      & 69.3 & -      & 75.9 & -      & 80.1 \\
    DenseNet-201    & 45.6 & 79.4 & 51.1 & 84.2 & 47.6 & 92.1 \\
    EfficientNet    & 37.9 & 73.3 & 43.4 & 76.1 & 35.1 & 87.5 \\
    GoogLeNet       & 45.1 & 70.8 & 48.8 & 75.8 & 38.9 & 84.1 \\
    WideResNet      & 49.3 & 78.1 & 53.0 & 80.9 & 37.8 & 89.8 \\
    VGG-16          & 55.5 & 71.5 & 58.8 & 74.1 & 49.9 & 86.1 \\
    GCN             & - & 72.9 & - & 77.3 & - & 88.4\\
    ViT             & 59.4 &  79.1 & 60.8 & 84.9 & 61.5 & 92.7\\
    \gv             & - & 86.9 &  - & 92.6 &  - & \cellcolor[HTML]{C0C0C0} 98.1 \\
\bottomrule
\end{tabular}
\end{table*}

Table \ref{tb:ClassificationWithoutProcessing} summarizes the gesture recognition success rate over the independent test set. The success rate refers to the ordinary accuracy of the model. The table reports the success rate for using the model in the following variations: raw images without pre-processing where the images are directly passed through the classifier; images after cropping-out the user using YOLOv3; and, with cropping-out and quality improvement using HQ-Net. First, fine-tuned models are shown to provide poor results as they were pre-trained to focus on features irrelevant to our specified problem. Training the models from scratch is much more beneficial and focuses them on the required task. Next, when comparing results with and without pre-processing, it is clear for all models that cropping-out the user from the image and improving quality with \hq significantly increase the success rate. Furthermore, GCN and ViT individually yield roughly comparable accuracy to the other benchmarked models. On the other hand, their combination in \gv provides superior accuracy. For \gv\!\!, cropping-out the images improves success rate over all classification models by $6.5\%$ on average. Adding also quality improvement provides an additional $5.9\%$ success rate increase. Overall, the proposed \gv model with cropping and quality improvement provides a superior and high recognition success rate over all models.  

To justify the use of \hq for image improvement over other methods evaluated in the previous section, we next evaluate the success rate of \gv with these methods. Hence, Table \ref{tb:Different_IQ} presents the success rate of \gv over the test data while using the quality improvement of the models evaluated in Section \ref{sec:IQI_eval}. Here also, having the \hq over the other methods is proven to be significantly superior. The proposed \hq model provides finer hand details and gestures can be recognized more clearly in the image. Note also that \gv without any SR provides higher accuracy than using any of the existing methods. This emphasizes the incompatibility of these methods to the ultra-range problem and the merits of \gv. In particular, an examination of Figures \ref{fig:SR9m}-\ref{fig:SR25m} reveals notable distinctions among the performance outcomes. These distinctions manifest in the manner by which the models handle image blurring where some induce uniform blurring across the entire image and ones that selectively obscure finer details. Upon closer inspection, it becomes evident that a significant proportion of the evaluated models tend to introduce image corruptions characterized by the degradation of crucial edge details. Consequently, such degradation adversely impacts the gesture recognition success rates with the other models, resulting in diminished overall performance metrics. On the other hand, the merits of combining \gv with \hq in achieving a high success rate in URGR is validated. Hence, these will be used in further evaluations.

\begin{figure}
    \centering
     \includegraphics[width=0.8\linewidth]{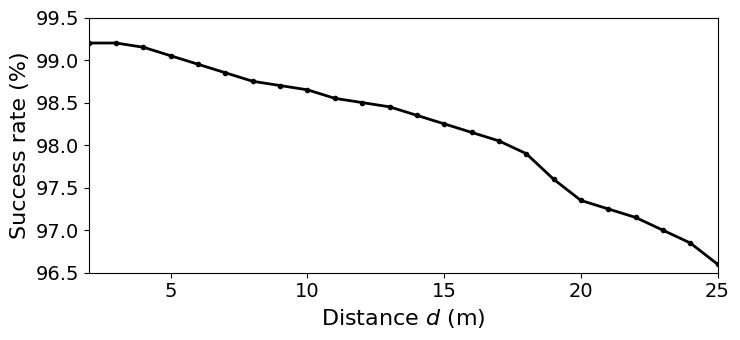} 
    \caption{Gesture recognition success rate of \gv with regards to the distance $d$ from the camera.}
    \label{fig:accuracy_vs_distance}
\end{figure}

\begin{figure}
    \centering
    \begin{tabular}{cc}
     \includegraphics[width=0.5\linewidth]{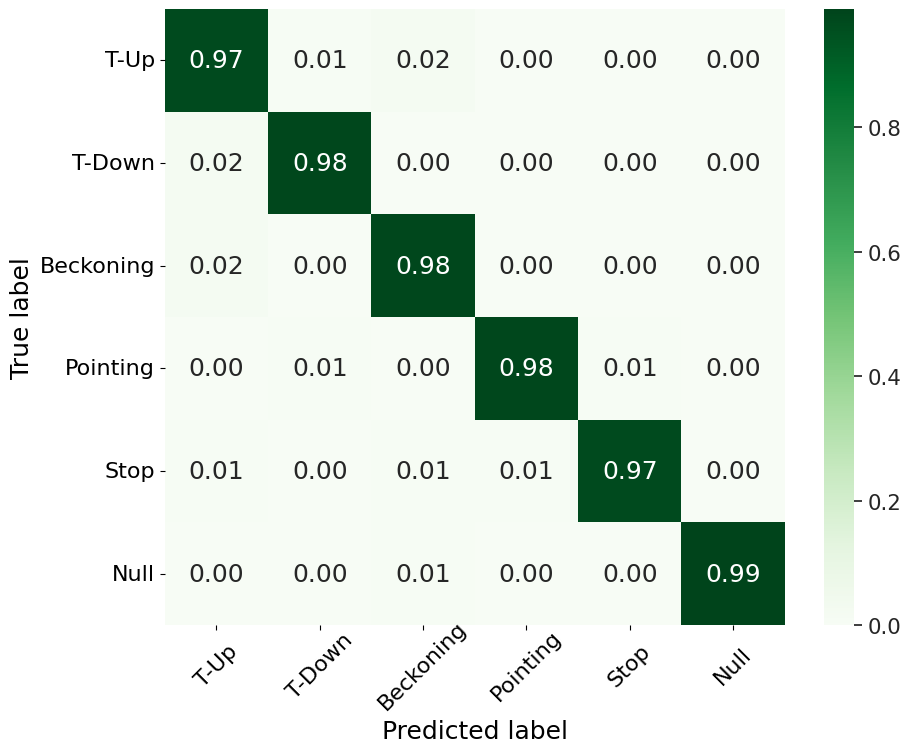} &
     \includegraphics[width=0.5\linewidth]{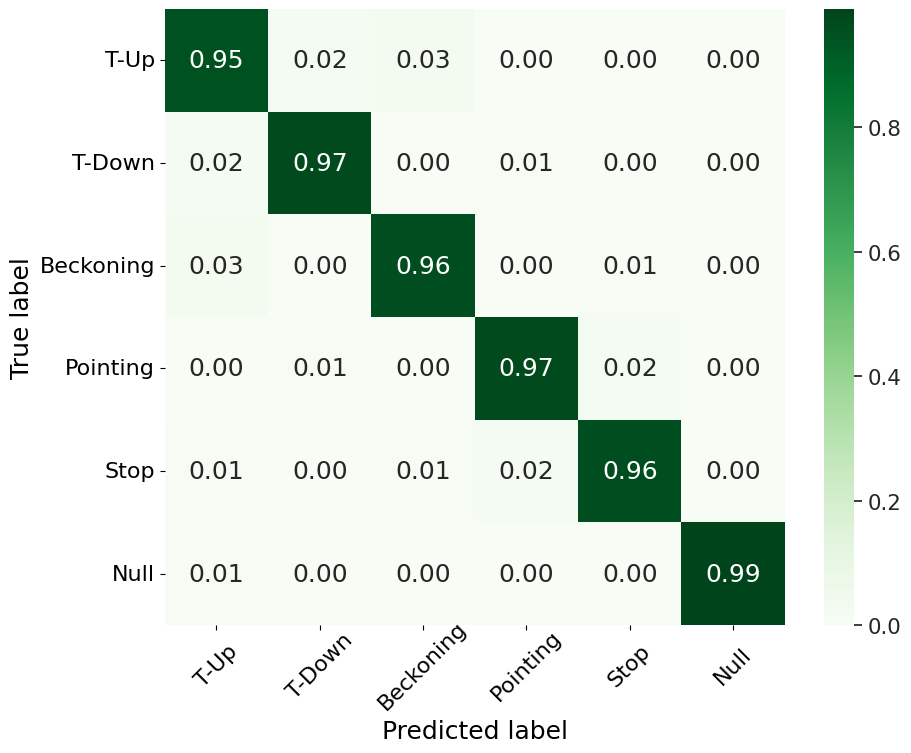} \\
     (a) & (b)
     \end{tabular}
    \caption{Classification confusion matrix for URGR with \gv over test data in the ranges (a) 15-20 meters and (b) 21-25 meters.}
    \label{fig:confmat}
\end{figure}

\begin{figure*}
    \centering
    \begin{tabular}{cccc} \hspace{-1.3cm}
        \includegraphics[height=0.23\linewidth]{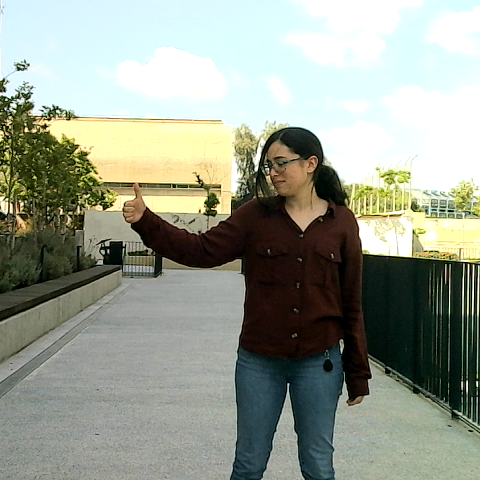} & 
        \includegraphics[height=0.23\linewidth]{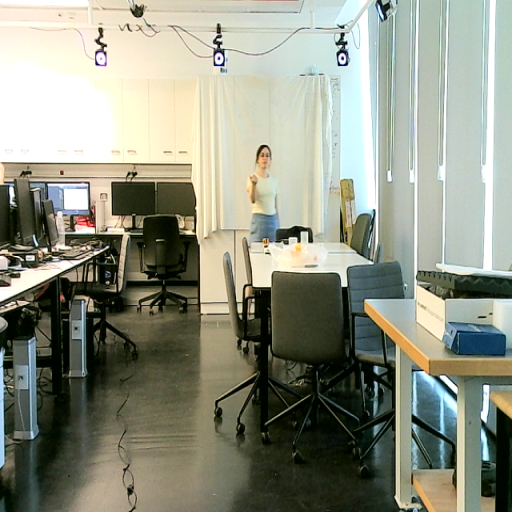} &
        \includegraphics[height=0.23\linewidth]{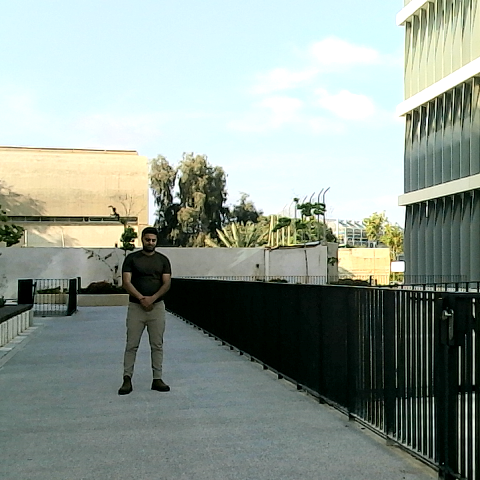} & 
        \includegraphics[height=0.23\linewidth]{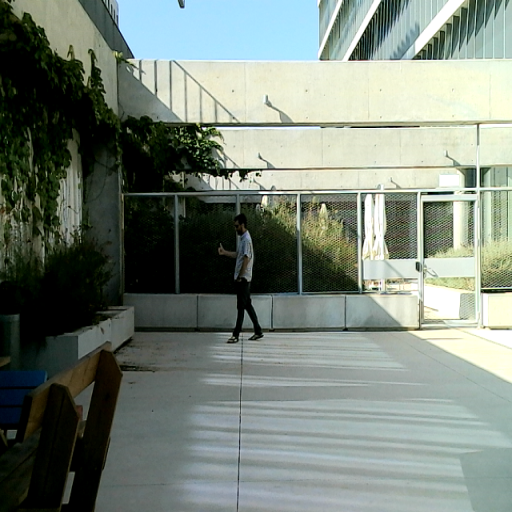} \\ \hspace{-1.3cm}
        Thumbs-up, $4~m$, 98.8\%  & 
        Beckoning, $9~m$, 98.6\% & 
        Null, $11~m$, 98.3\% & 
        Thumbs-up, $13~m$, 98.2\% \\ \hspace{-1.3cm}
        \includegraphics[height=0.23\linewidth]{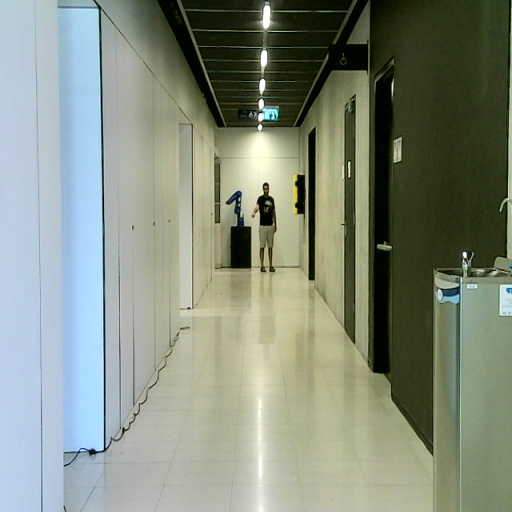} &
        \includegraphics[height=0.23\linewidth]{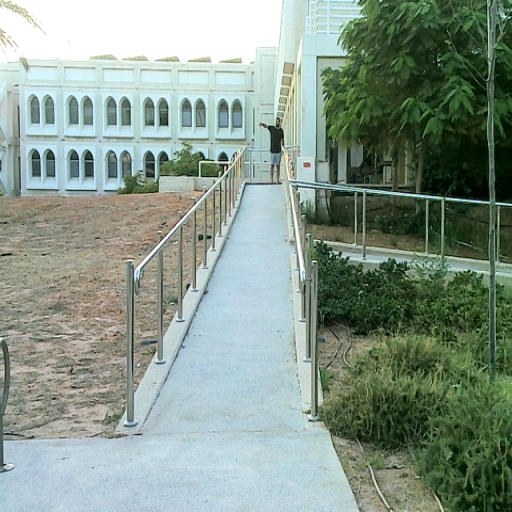} & \includegraphics[height=0.23\linewidth]{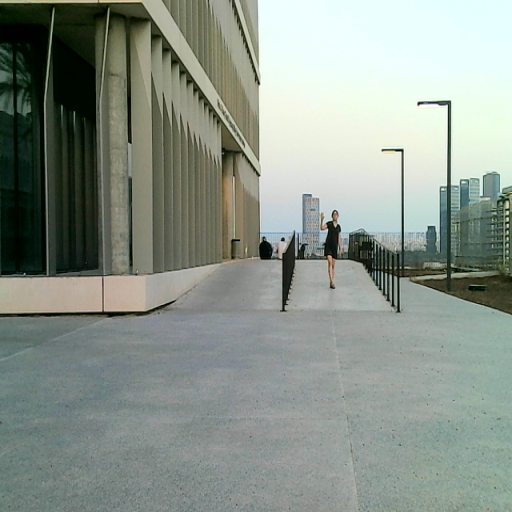} & \includegraphics[height=0.23\linewidth]{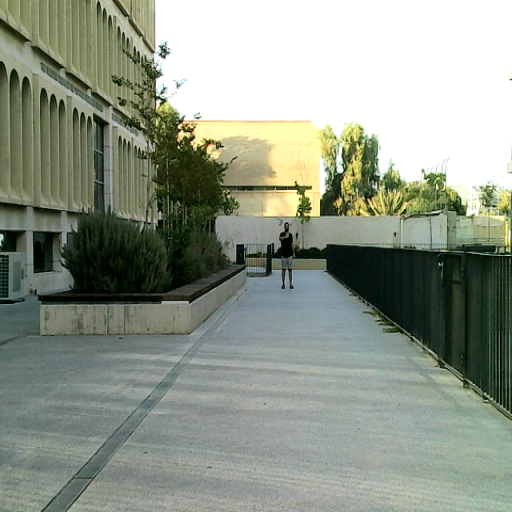} \\ \hspace{-1.3cm}
        Pointing, $18~m$, 97.5\% & Thumbs-down, $20~m$, 97.3\% & Stop, $23~m$, 96.8\% & Thumbs-up, $25~m$, 98.8\% \\
    \end{tabular}
    \caption{Examples of correct gesture recognition with GViT. Each snapshot is denoted by the exhibited gesture, its distance $d$ and the model certainty.}
    \label{fig:dis_exm}
\end{figure*}

\begin{figure*}
    \centering
    \begin{tabular}{cccc} \hspace{-1.3cm} 
        \includegraphics[height=0.195\linewidth]{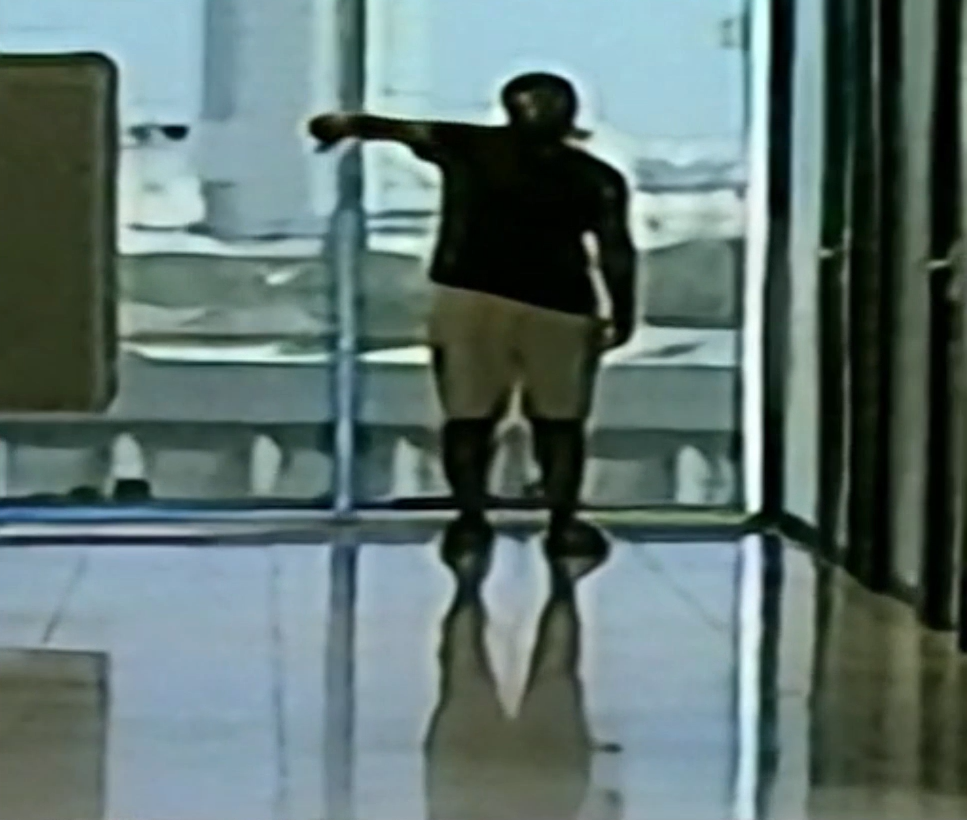} & \hspace{-0.3cm} 
        \includegraphics[height=0.195\linewidth]{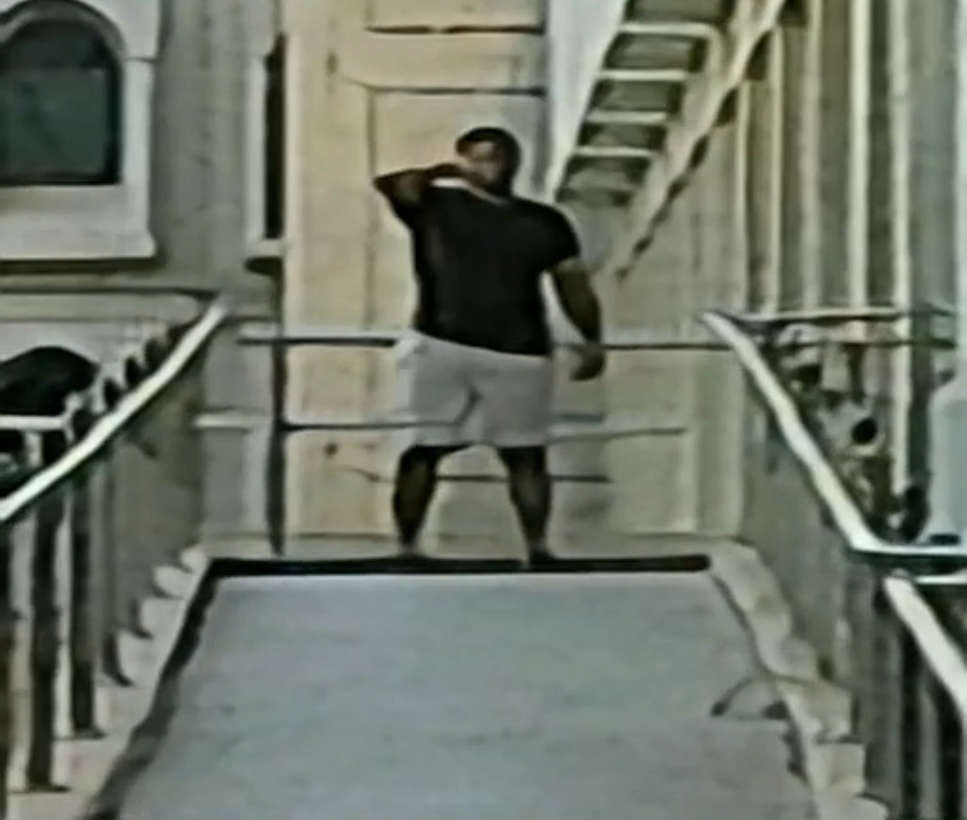} & \hspace{-0.3cm} 
        \includegraphics[height=0.195\linewidth]{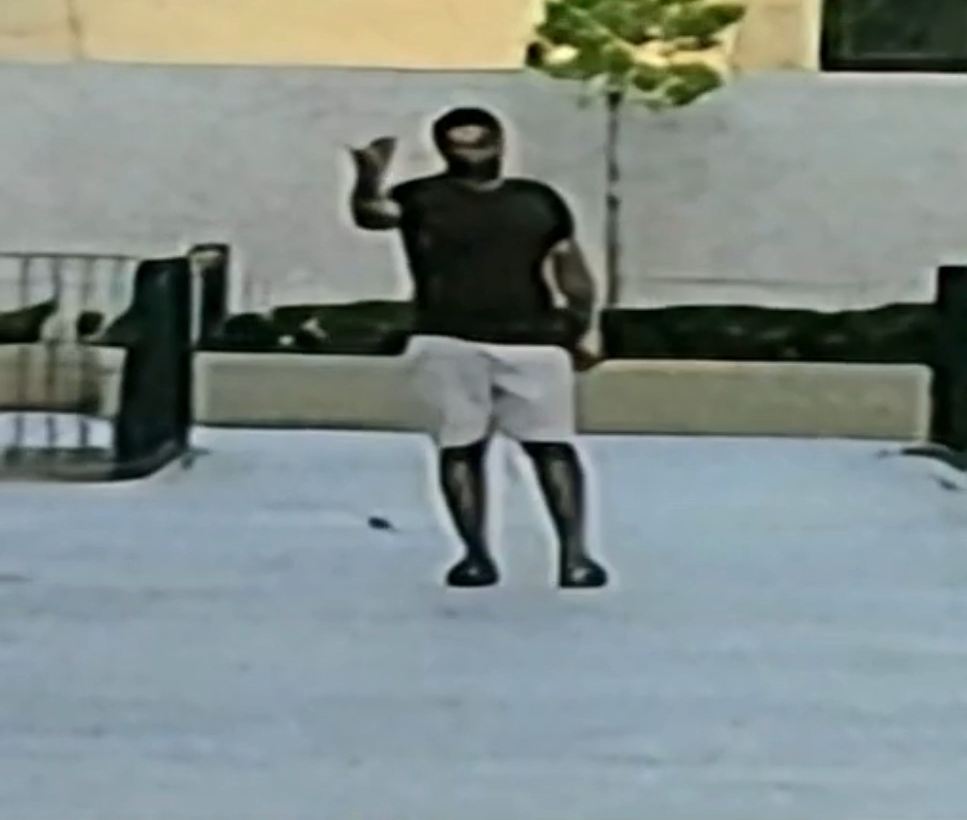} & \hspace{-0.3cm} 
        \includegraphics[height=0.195\linewidth]{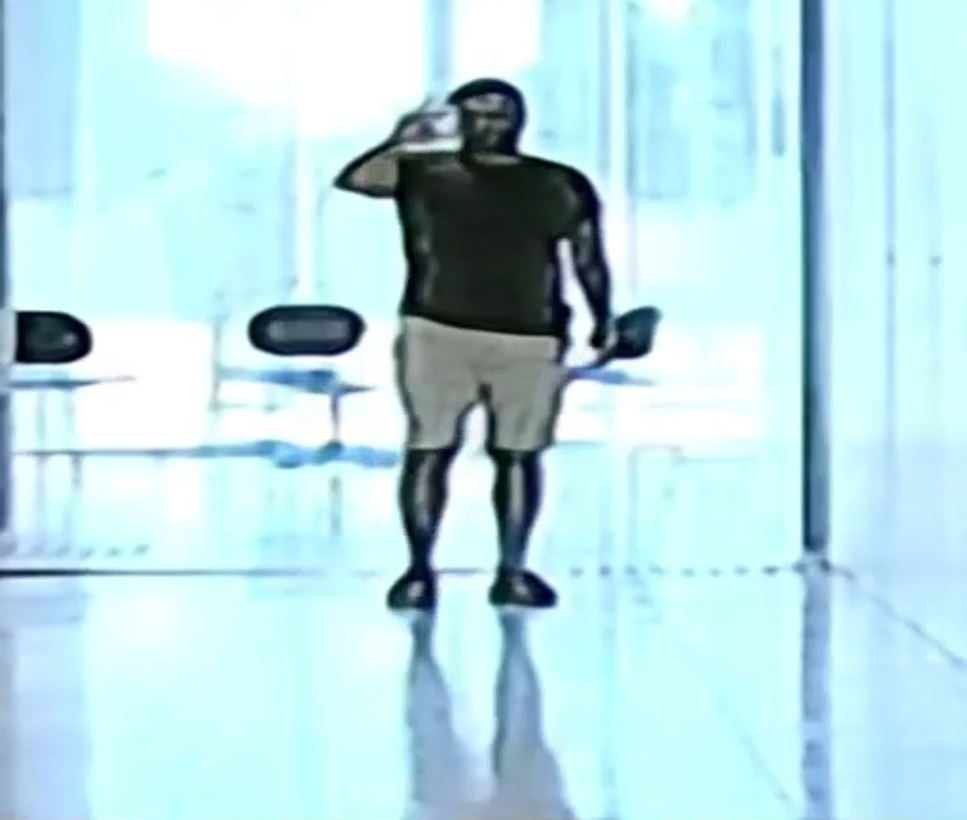} \\ \hspace{-1.3cm} 
        Thumbs-down, $20~m$, 90.0\% & \hspace{-0.3cm} 
        Thumbs-down, $25~m$, 90.3\% & \hspace{-0.3cm} 
        Beckoning, $25~m$, 88.3\% & \hspace{-0.3cm} 
        Thumbs-up, $23~m$, 84.2\% \\ \hspace{-1.3cm} 
        \includegraphics[height=0.195\linewidth]{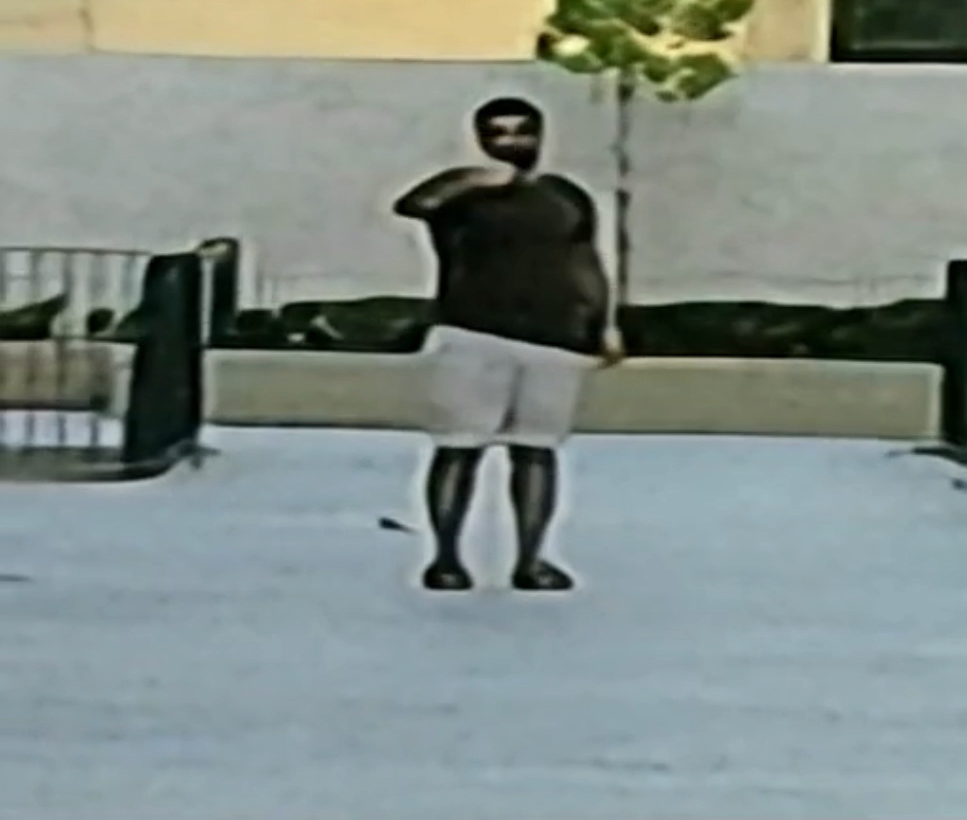} & \hspace{-0.3cm} 
        \includegraphics[height=0.195\linewidth]{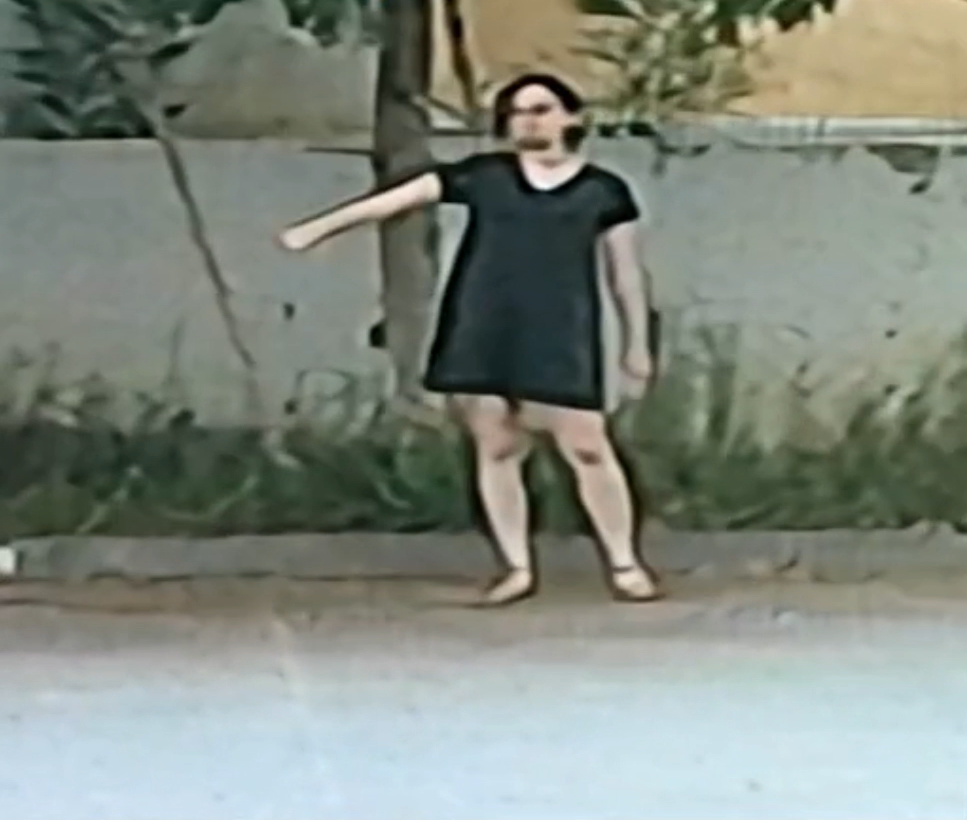} & \hspace{-0.3cm} \includegraphics[height=0.195\linewidth]{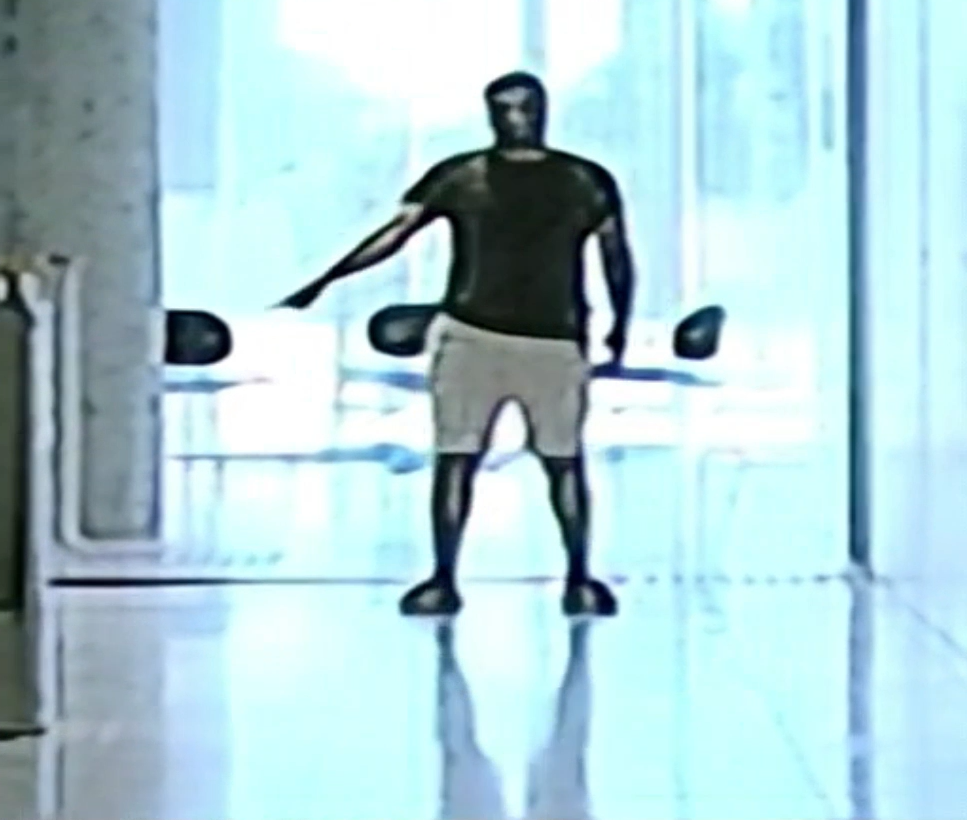} & \hspace{-0.3cm} \includegraphics[height=0.195\linewidth]{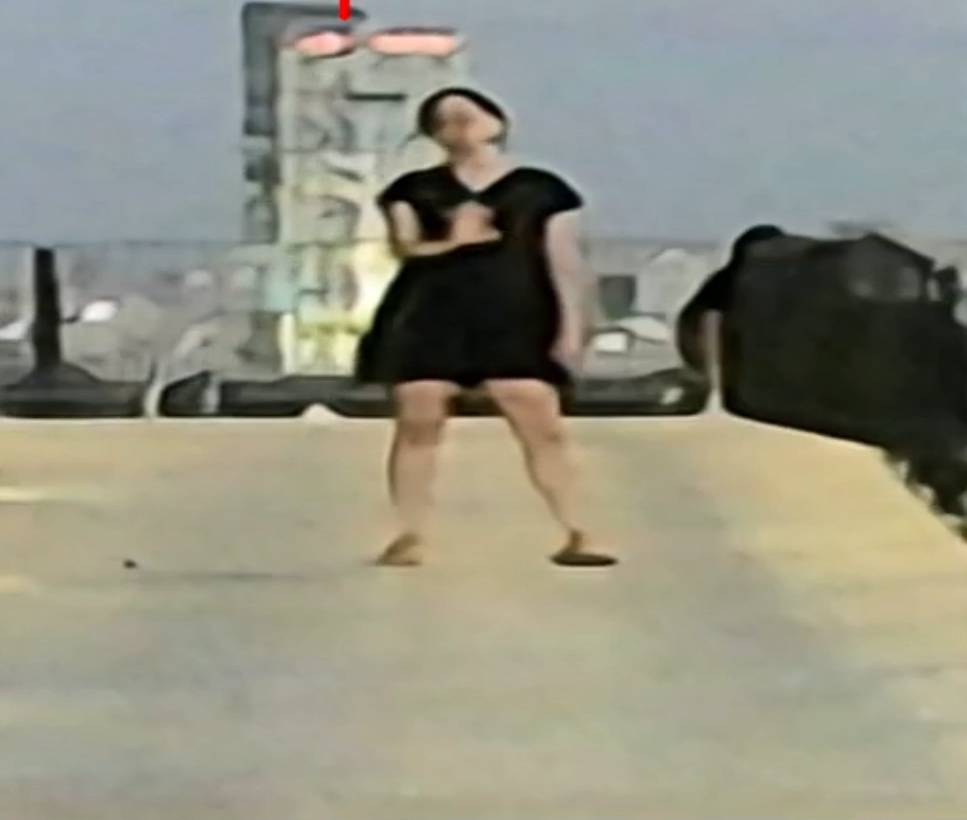} \\ \hspace{-1.3cm} 
        Thumbs-up, $25~m$, 87.6\% & \hspace{-0.3cm} 
        Pointing, $25~m$, 87.6\% & \hspace{-0.3cm} 
        Pointing, $23~m$, 89.1\% & \hspace{-0.3cm} 
        Stop, $20~m$, 81.0\% \\
    \end{tabular}
    \caption{Examples of correct gesture recognition in real-time with GViT. Images are shown after focusing on the user. Each snapshot is denoted by the exhibited gesture, its distance $d$ and the model certainty.}
    \label{fig:dis_exm_close}
\end{figure*}

In the next analysis, seen in Figure \ref{fig:accuracy_vs_distance}, the recognition success rate of \gv is evaluated with regards to the distance $d$ from the camera. Each point along the curve is the percentage of successful trials out of 1,000 recognition attempts in the corresponding distance. In short ranges, the success rate is approximately 99\% while in ultra-range it is slightly reduced to 96.6\% at 25 meters. Figure \ref{fig:confmat} shows the confusion matrices over test data of distance ranges 15-20 meters and 21-25 meters. These results showcase the high efficiency of the \gv model throughout the entire range of work. Figure \ref{fig:dis_exm} exhibits various examples of gesture recognition along with model certainty in different indoor and outdoor environments. Model certainty is the probability $P(\mathcal{O}_j|\hve{I}_i)$ of image $\hve{I}_i$ to be classified to $\mathcal{O}_j$ and is acquired by the output of the softmax layer. We further analyze the performance of the URGR framework in real-time. The framework was deployed in real-time while measuring the average inference frequency, yielding a frequency of 11.43 Hz. Furthermore, Figure \ref{fig:dis_exm_close} shows snapshot examples of gesture recognition when rolling-out the framework in real-time. \ref{apdx:failure} includes some examples of failures in gesture recognition. The presented results validate and showcase the high performance of the proposed URGR framework in gesture recognition in various ultra-range environments and its ability to work in real-time.

\begin{table}
\centering
\caption{Gesture recognition success rate using \gv while improving image quality with various SR models}
\label{tb:Different_IQ}
\begin{tabular}{lc}\toprule
        Models~~~~~~~~~~~~~~~~~~          & Success rate (\%)  \\\midrule
        AutoEncoder     & 80.3 \\
        Unet            & 82.6 \\
        Unet++          & 84.2 \\
        ESRGAN          & 86.5 \\
        RUNet           & 87.1 \\
        BSRGAN          & 89.3 \\
        No SR           & 92.6 \\
        \rowcolor{lightgray}
        \hq             & 98.1 \\
\bottomrule
\end{tabular}
\end{table}

\subsection{Data requirements}

\begin{figure}
    \centering
     \includegraphics[width=\linewidth]{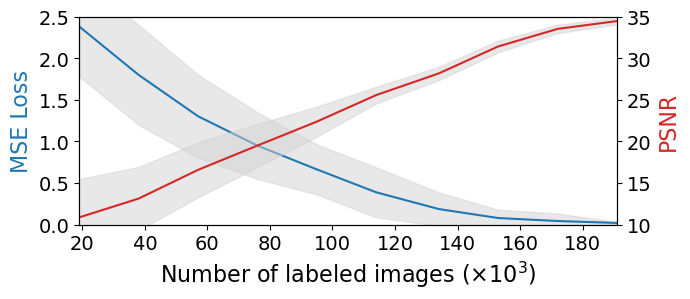} 
    \caption{Image quality improvement of the \hq model evaluated with the MSE loss and PSNR over the test data and with regards to the amount of data used to train the model.}
    \label{fig:HQ-Net_data}
\end{figure}

An evaluation of the data requirements for training \hq and \gv is given next. First, we evaluate the MSE loss and PSNR of \hq with the increase in data and up to $M=191,574$. For a specific amount of data, the model was cross validated over five batches taken randomly from the entire set $\mathcal{S}$. Figure \ref{fig:HQ-Net_data} presents the mean MSE loss and PSNR of \hq over the test data with regard to the size of the training data. The results show constant improvement with the increase of training data until reaching some saturation with over $180,000$ images.

\begin{figure}
    \centering
    \includegraphics[width=\linewidth]{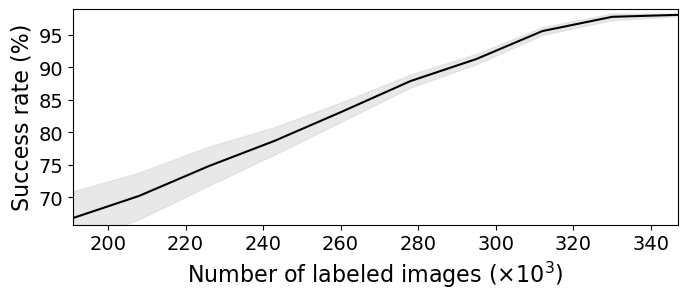} 
    \caption{Gesture recognition success rate of \gv over the test data with regards to the amount of data used to train the model.}
    \label{fig:accuracy_GVIT_data}
\end{figure}

With the fully trained \hq model, we next evaluate the \gv model for the required amount of data. Since the training of \hq has already used $M$ labeled images in the range of $[2,8]$ meters, these available images are also used to train \gv\!\!. Therefore, we begin the evaluation with the existing $M$ images of up to 8 meters. Similar to \hq\!, the \gv model is cross-validated, for each data size, over five batches taken randomly from $\mathcal{H}$. Figure \ref{fig:accuracy_GVIT_data} shows the success rate of gesture recognition over the test data with regards to the amount of data and up to $N=347,483$. The results show that training with data of up to 8 meters is not sufficient for successful recognition in ultra-range. Adding diverse images from all ranges is shown to improve accuracy and reach up to 98.1\% success rate with $N$ labeled images.

\subsection{Edge cases}

The training data in $\mathcal{H}$ does not include edge case scenarios which may limit performance. First, as described above, the training data was collected using only the right hand of the participants. Hence, we have collected an additional test set of 6,000 labeled images taken from four participants while only gesturing with their left hand and in various ultra-range environments. Using the models trained on dataset $\mathcal{H}$ and presented in Table \ref{tb:ClassificationWithoutProcessing}, the new test set was evaluated. The success rates for these models including \gv are seen in Table \ref{tb:LeftHand}. Here also, \gv dominates all other models with a high success rate of 97\%. In addition, there is a slight but marginal decline compared to the right arm. The results distinctly highlight the ability of the trained \gv to generalize to human gestures executed with the left hand.

\begin{table}
\centering
\caption{Comparison of gesture recognition success rates with the various trained models on test data with only the left hand. For comparison, results from the right arm are also included.}
\label{tb:LeftHand}
\begin{tabular}{lcc}\toprule 
             & \multicolumn{2}{c}{Success rate (\%)} \\
    Models          & ~~~~~~~~~~Left arm~~~~~~~~~~ & ~~~~~~~~~~(Right arm)~~~~~~~~~~  \\\midrule
    CNN             & 75.8 & 80.1 \\
    DenseNet        & 89.4 & 92.1 \\
    EfficientNet    & 86.2 & 87.5 \\
    GoogLeNet       & 80.7 & 84.1 \\
    WideResNet      & 88.4 & 89.8 \\
    VGG-16          & 79.3 & 86.1 \\
    GCN             & 81.1 & 84.4 \\
    ViT             & 88.9 & 92.7 \\
    \rowcolor{lightgray}
    \gv             & 97.0 & 98.1 \\
    \bottomrule
\end{tabular}
\end{table}

\begin{figure*}
    \centering
    \begin{tabular}{cccc}
        \includegraphics[height=0.23\linewidth]{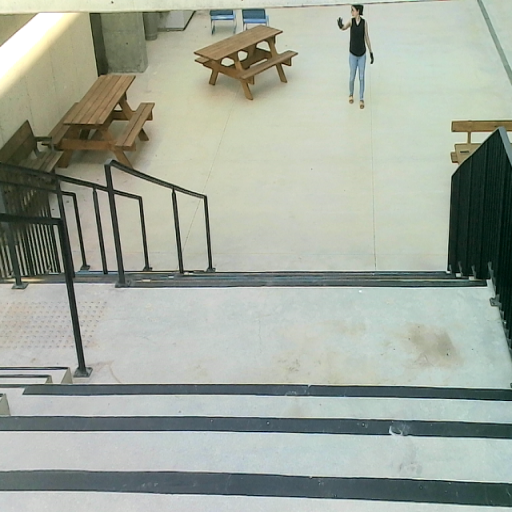} & 
        \includegraphics[height=0.23\linewidth]{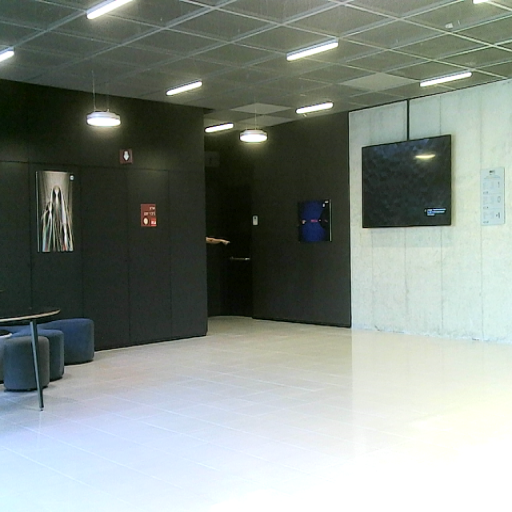} &
        \includegraphics[height=0.23\linewidth]{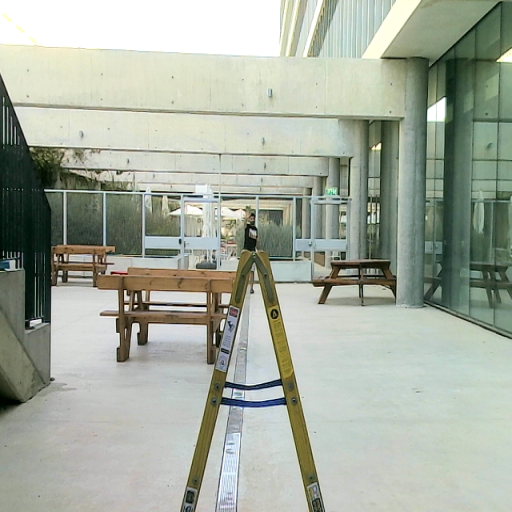} & 
        \includegraphics[height=0.23\linewidth]{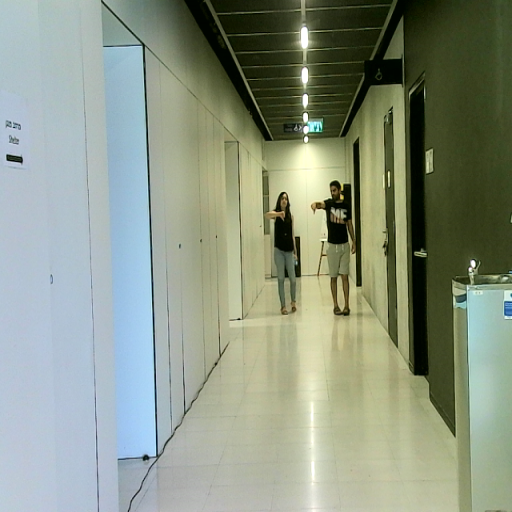} \\
        (a) & (b) & (c) & (d) \\
        \includegraphics[height=0.23\linewidth]{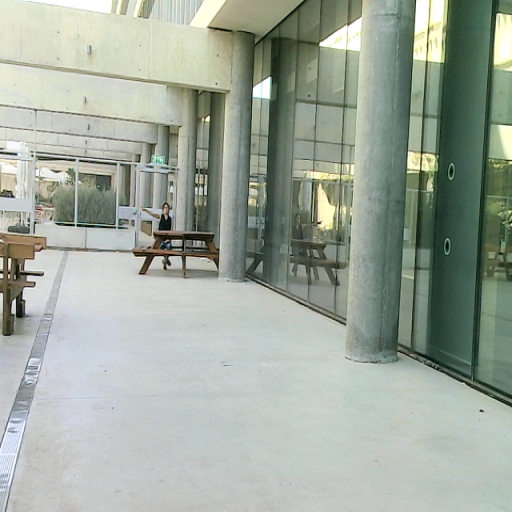} &
        \includegraphics[height=0.23\linewidth]{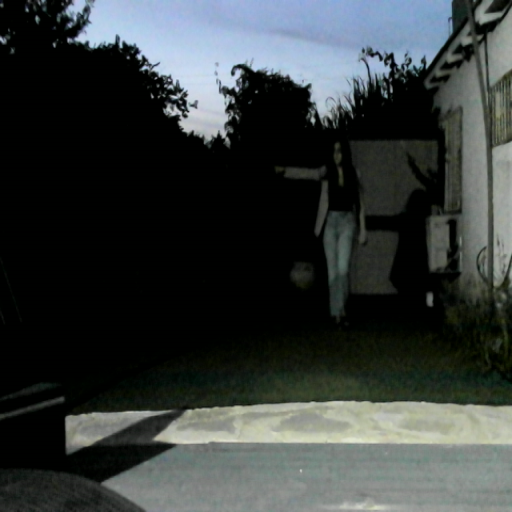} & \includegraphics[height=0.23\linewidth]{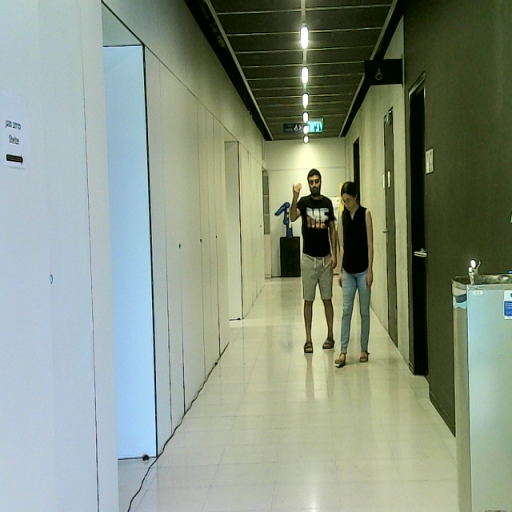} & \includegraphics[height=0.23\linewidth]{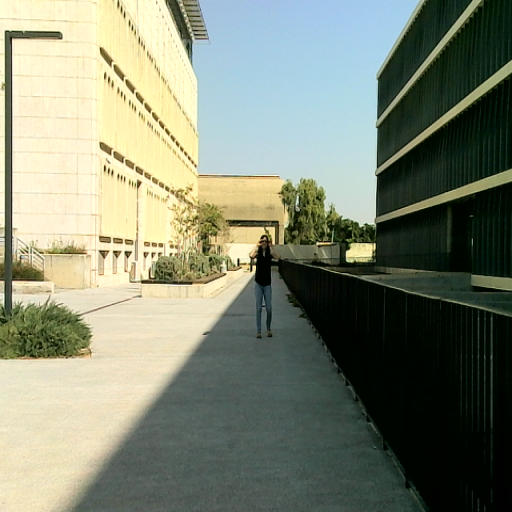} \\
        (e) & (f) & (g) & (h) \\
    \end{tabular}
    \caption{Examples of correct gesture recognition with \gv in several edge cases not included in the training data: (a) stop gesture from 17 meters distance while wearing gloves, having model certainty of 94.3\%; (b) pointing gesture from 13 meters distance while the participant is fully occluded, having model certainty of 95.8\%; (c) stop gesture from 22 meters distance while the participant is partly occluded by a ladder, having model certainty of 90.1\%; (d) thumbs-down gesture from 14 meters distance from two participants, having model certainty of 93.9\%. (e) pointing gesture from 23 meters distance while the participant is sitting down, having model certainty of 94.5\%;  (f) pointing gesture from 11 meters distance in a poor lighting environment, having model certainty of 88.2\%; (g) beckoning gesture from 12 meters distance while having another null participant interfering in the foreground, having model certainty of 91.1\%; and (h) null gesture from 19 meters distance while the participant is lifting both hands, having model certainty of 87.7\%.}
    \label{fig:edge}
\end{figure*}

Furthermore, we evaluate specific but interesting edge cases where accurate recognition may be difficult. Designated test sets were collected for eight edge cases including: gloved hands; out-of-frame participant where only the arm is visible; the participant is partly occluded by an object in the foreground; gesturing while the participant is seated; dual-arm lift while only one arm is exhibiting a gesture; environment with poor lighting; multiple participants in the image exhibiting the same gesture; and interference by another person in the foreground. Such edge cases were not included in the training. A test set of 1,000 labeled images for each edge case was collected in various indoor and outdoor environments. Table \ref{tb:edge} presents the gesture recognition success rates for these cases with GViT. The majority of the cases achieved high success rates. Nevertheless, dual-arm lift and poor lighting yielded slightly lower results which could be mitigated by additional data in such scenarios. Figure \ref{fig:edge} exhibits various examples of gesture recognition in different edge cases along with model certainty in different indoor and outdoor environments. \ref{apdx:failure} includes some examples of gesture recognition failures in several edge cases. Overall, \gv is able to generalize and recognize gestures in unconventional and challenging scenarios.

\begin{table}[]
\centering
\caption{Gesture recognition success rate for several edge cases}
\label{tb:edge}
\begin{tabular}{lc} \toprule
    Edge Cases              & Success rate (\%)  \\\midrule
    Gloves                  & 95.7  \\
    Out-of-frame            & 96.1 \\
    Occlusions              & 91.8 \\
    Sitting-down            & 95.9 \\
    Dual-arm lift           & 88.7 \\
    Poor lighting           & 89.5 \\
    Multiple participants   & 94.1\\
    Interference            & 91.6\\
    \bottomrule
\end{tabular}
\end{table}

\subsection{Robot Experiments}
\label{sec:experiments}

\begin{figure}
    \centering
    \includegraphics[width=0.55\linewidth]{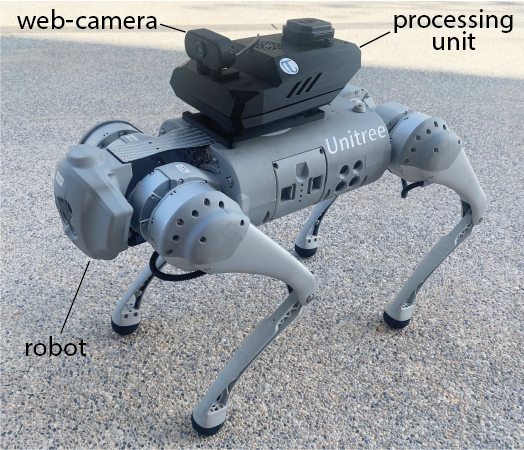} 
    \caption{Robotic platform used in the experiment based on the Unitree Go1 quadruped robot equipped with a simple RGB web-camera.}
    \label{fig:robot}
\end{figure}

Following the training and evaluation of \gv\!\!, the proposed URGR framework is now demonstrated and evaluated in ultra-range HRI. In the experiment, we wish to show real-time responses of a robot to recognized gestures exhibited by a user positioned in ultra-range. Hence, we have set up a mobile robotic platform based on the Unitree Go1 quadruped robot seen in Figure \ref{fig:robot}. A processing unit with a simple RGB web-camera was mounted on the robot. The processing unit is based on an Nvidia Jetson Orin Nano connected to the robot with the Unitree high-level Robot Operating System (ROS) API. Videos of the experiments can be seen in the supplementary material. While the frequency of the URGR framework was previously shown to be 11.43 Hz, a communication latency has resulted in visible delays between the gesture and response. The latency is a result of having the computation on a remote computer while acquired images are wirelessly transferred at a low frequency. Future work should have the URGR framework run onboard the robot for better performance.

The processing unit runs the URGR framework in real-time and, upon recognition of a gesture, commands the robot to move to comply. For evaluation and demonstration, we set a premeditated response to each of the six gestures in Figure \ref{fig:gestures}: a pointing gesture commands the robot to move sideways; thumbs-up will instruct the robot to pitch tilt; thumbs-down will make the robot lie down; beckoning commands the robot to move toward the user; stop gesture will make the robot halt any previous command; and upon null, the robot will not change its current behavior (e.g., moving forward). In the case of the robot moving sideways, the camera may lose sight of the user in the frame. In our case, the user will also move to remain in the frame. In future work, one can consider an algorithm for which the robot will act to maintain a line-of-sight with the user.

\begin{figure}
    \centering
    \includegraphics[height=0.3\textheight]{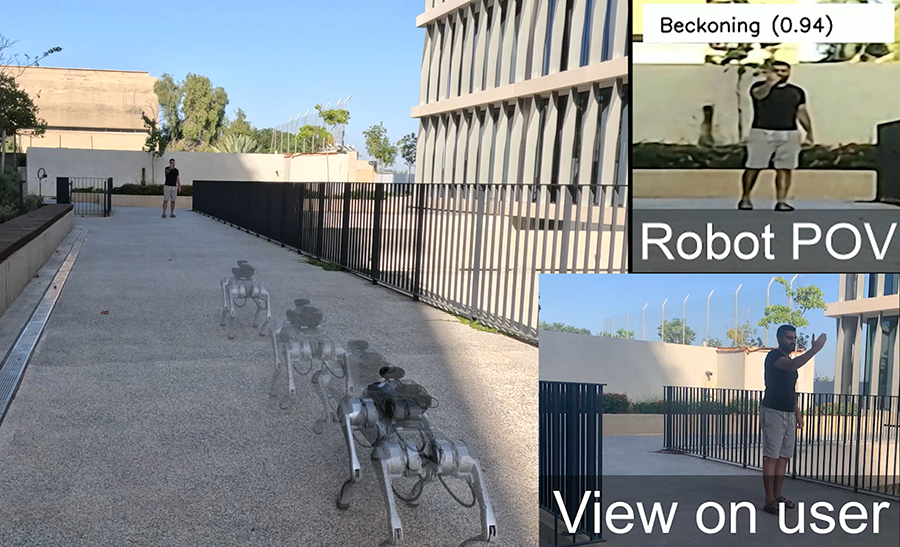} \\
    \vspace{0.1cm}
    \includegraphics[height=0.3\textheight]{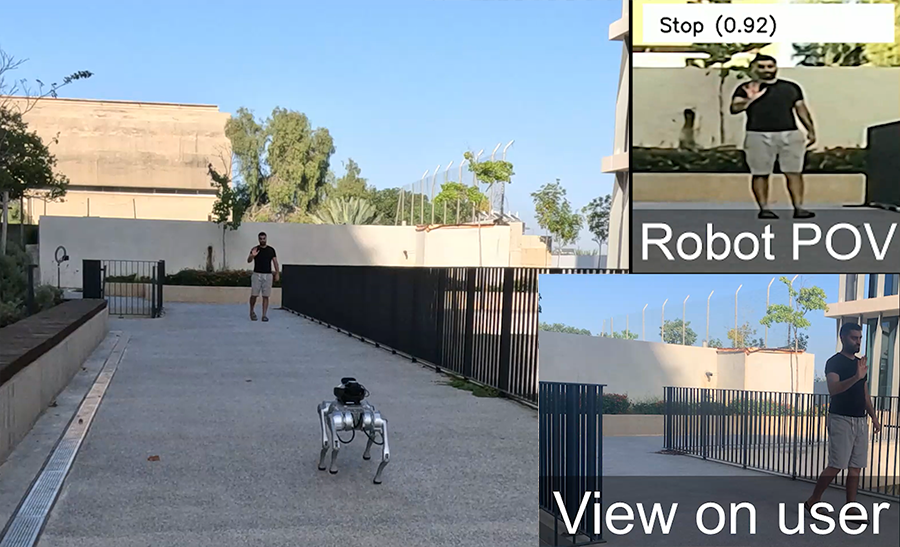} \\
    \vspace{0.1cm}
    \includegraphics[height=0.3\textheight]{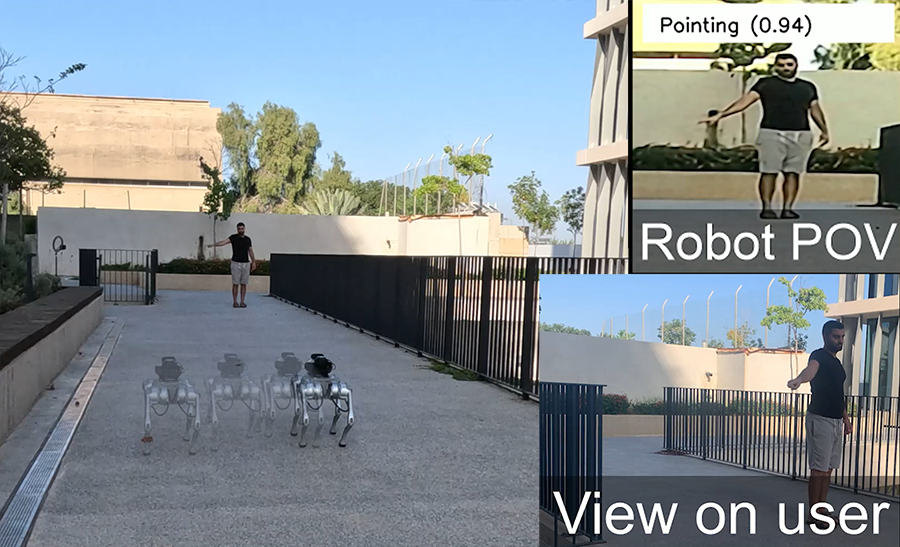}
    \caption{A user is directing a robot in ultra-range and in an outdoor environment: (top) Starting at 25 meters, the user is exhibiting a beckoning gesture leading to the robot moving forward; (middle) the user is exhibiting a stop gesture making the robot halt; and, (bottom) the user is pointing so to make the robot move sideways.}
    \label{fig:exp_30}
\end{figure}

\begin{figure}
    \centering
    \includegraphics[width=\linewidth]{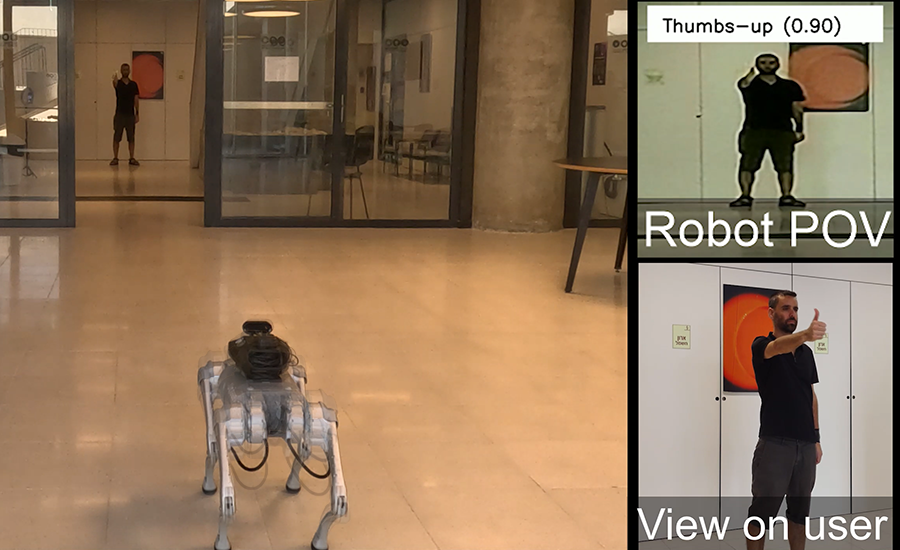} \\
    \vspace{0.1cm}
    \includegraphics[width=\linewidth]{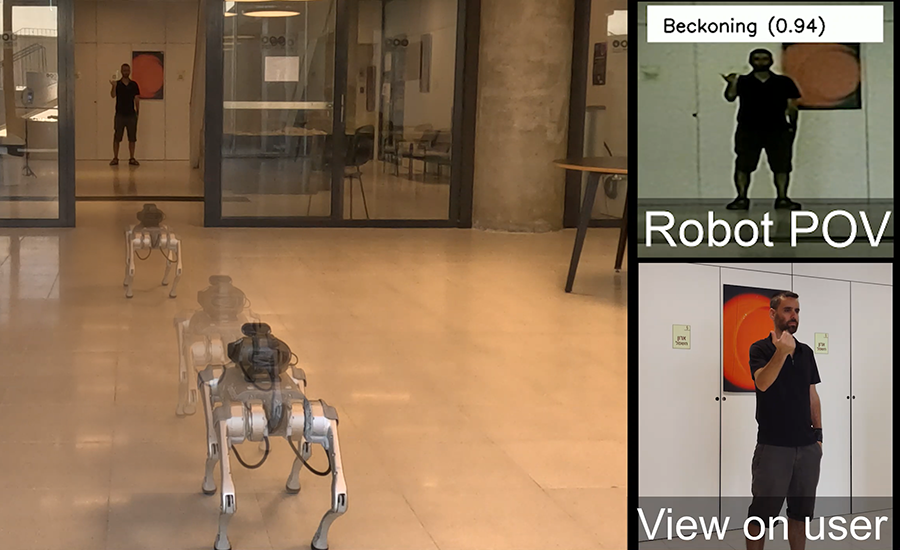} \\
    \caption{A user is directing a robot from a distance of 21 meters in an indoor environment and through a door opening: (top) The user is exhibiting a thumbs-up gesture leading to the robot tilting; and, (bottom) the user is exhibiting a beckoning gesture leading to the robot moving forward.}
    \label{fig:exp_17}
\end{figure}

\begin{figure}
    \centering
    \includegraphics[width=\linewidth]{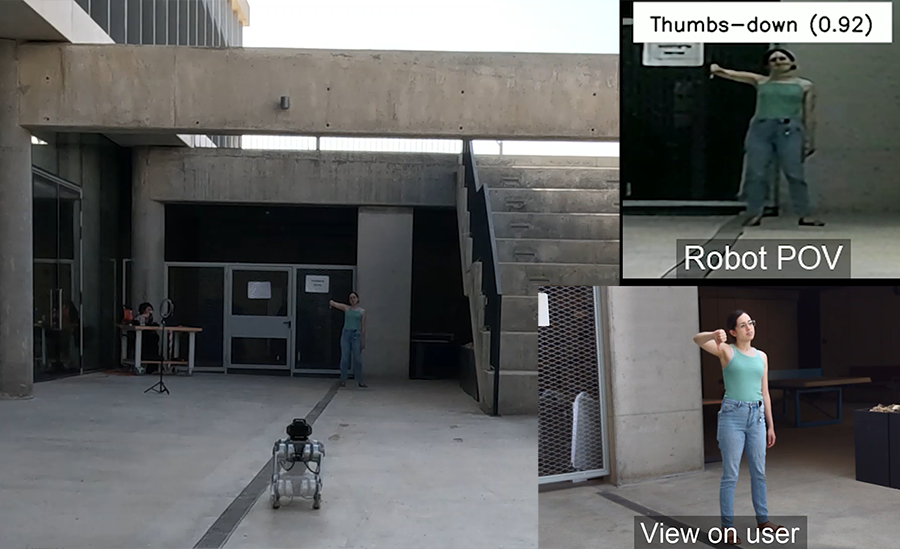} 
    \caption{A user is directing a robot from a distance of 15 meters in a courtyard environment. The user is exhibiting a thumbs-down gesture leading the robot to lie down.}
    \label{fig:exp_11}
\end{figure}

We observe the gesture commands and these robot responses in various distance ranges and in three environments: outdoor, indoor and in a courtyard. Within each environment, 100 gesture trials were exhibited. Table \ref{tb:Real-time} reports the success rates of acquiring the correct response from the robot. Figures \ref{fig:exp_30}-\ref{fig:exp_11} show snapshots of some gesture responses of the robot in the three environments. The snapshots include the robot's point-of-view (POV) and a closer view of the user for clear verification. While the work focuses on a distance range of up to 25 meters, we have included gestures in a distance of 28 meters and outdoors. The results show a high response rate to the directives of the user. Hence, the URGR framework is validated in real-time operation.

\begin{table}
\centering
\caption{Real-time gesture recognition success rate in different environments}
\label{tb:Real-time}
\begin{tabular}{lcc}\toprule
    Environment & Distance range (m) & Success rate (\%) \\\midrule
    Outdoor   & 15-28 & 98 \\ 
    Indoor    & 12-24 & 97 \\ 
    Courtyard & 17-20 & 94 \\
\bottomrule
\end{tabular}
\end{table}


\section{Conclusions}

In this work, an ultra-range gesture recognition framework has been introduced to facilitate a robot's ability to interpret human gestures in an ultra-range distance with an RGB camera and subsequently act accordingly. A pivotal component of this framework is the innovative \hq model, tailored to enhance the quality of distant objects within images. This is a critical aspect for ensuring successful human gesture recognition. On top of the low-resolution problem of the images due to zooming-in on the user, distant objects may experience image degradation. Hence, \hq is intended to enhance image details for further recognition. The second novel model, GViT, is a fusion of Graph Convolutional Network and Vision Transformers designed for the recognition of human hand gestures. Rigorous testing across a diverse test dataset has validated the superiority of \gv over state-of-the-art methodologies. Notably, \gv achieved a remarkable 98.1\% success rate of recognizing human gestures at distances of up to 25 meters. Furthermore, \hq was justified compared to other super resolution approaches. Our findings have also revealed that our proposed ultra-range gesture recognition framework outperforms human perception. The framework was deployed on a quadruped robot and evaluated its response to six gestures in ultra-range. The robot demonstrated high response rates even at a distance of up to 28 meters. 


Our approach has yielded a high success rate for recognizing human gestures. However, in some edge cases, the recognition rate declines due to a lack of corresponding training data. Future work could broaden the dataset to generalize better. Alternatively, simulated or generative synthetic data can be included. Our framework is based on recognition from a single image. Future work may also consider temporal inference such that the robot can have higher certainty on the gesture class. In addition, it would be able to recognize the gesture at early stages of its exhibition for seamless operation. While our proposed framework is focused on gesture recognition, future work may adapt it to other object recognition tasks in ultra-range. This may include surveillance, satellite imagery and sports. For general usage (e.g., sign language), future work may expand the approach for a large number gestures. In addition, subsequent research could explore the recognition of human gestures in challenging environmental conditions (such as bad weather or smoke) and over longer distances, with a particular focus on distances of up to 40 meters. To achieve such a goal, data collection is required in the longer distances. Therefore, future work may consider image data augmentation and generation based on existing close range data to learn ultra-range recognition. Consequently, synthetic data would be generated in the longer ranges with no additional collection of real data. Additionally, investigations into the feasibility of drones recognizing human gestures from a distance are promising avenues for future exploration. Furthermore, integration with verbal instructions may provide a complementary capability for seamless and context-aware communication. 



\appendix
\section{Failure examples}
\label{apdx:failure}

\begin{figure*}
    \centering
    \begin{tabular}{ccc} 
        \includegraphics[height=0.3\linewidth]{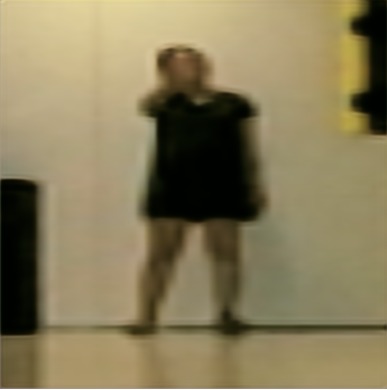} & 
        \includegraphics[height=0.3\linewidth]{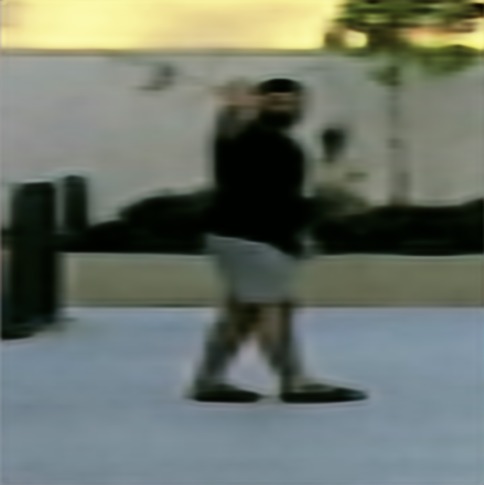} &
        \includegraphics[height=0.3\linewidth]{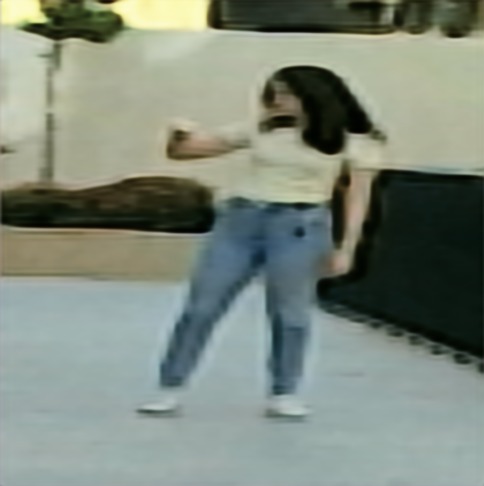} \\
        Thumbs-up, $24~m$, 48\%  & 
        Stop, $25~m$, 61\% & 
        Thumbs-down, $13~m$, 53\% \\ 
    \end{tabular}
    \caption{Examples of failed gesture recognition with GViT. Each snapshot is denoted by the wrongly predicted gesture, its distance $d$ and the model certainty. The true gestures are, from left to right, beckoning, thumbs-down and thumbs-up.}
    \label{fig:failure}
\end{figure*}

\begin{figure*}
    \centering
    \begin{tabular}{ccc} 
        \includegraphics[height=0.3\linewidth]{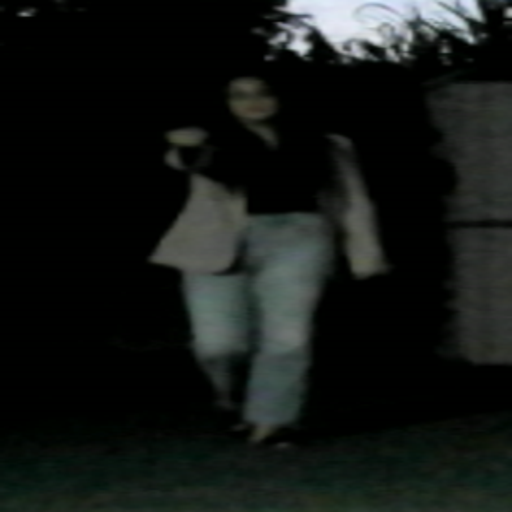} & 
        \includegraphics[height=0.3\linewidth]{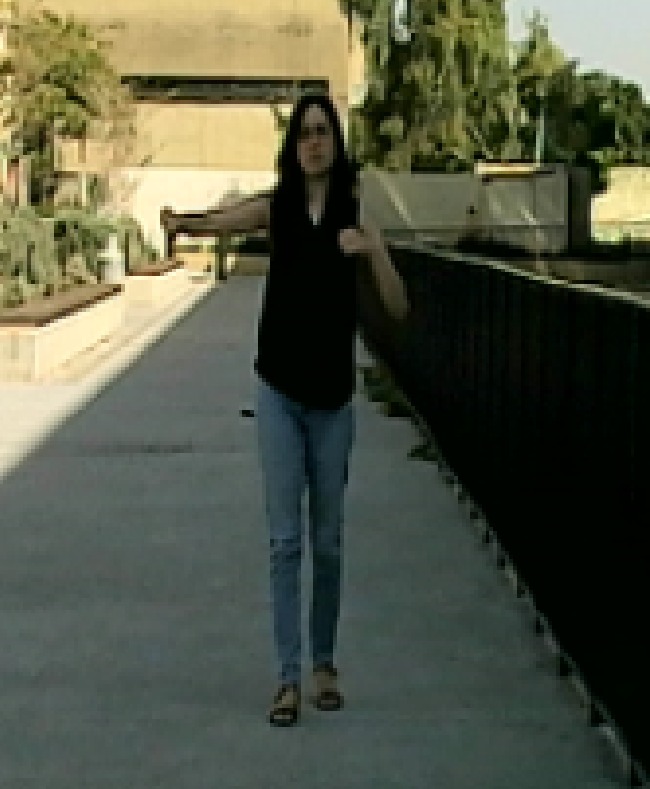} &
        \includegraphics[height=0.3\linewidth]{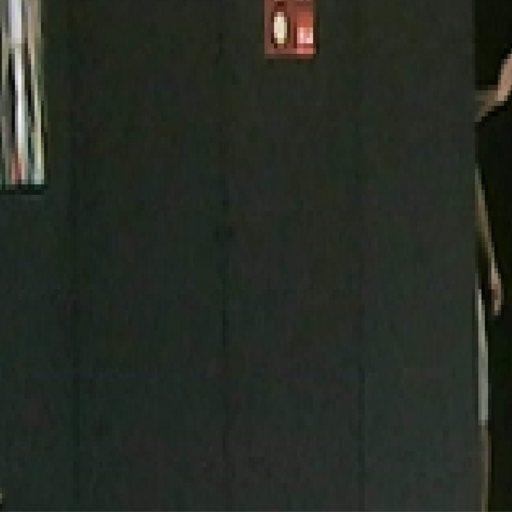} \\
        Beckoning, $24~m$, 51\%  & 
        Pointing, $15~m$, 43\% & 
        Stop, $18~m$, 35\% \\ 
    \end{tabular}
    \caption{Examples of failed gesture recognition with GViT in several edge cases. Each snapshot is denoted by the wrongly predicted gesture, its distance $d$ and the model certainty. From left to right, the true gestures are thumbs-up in poor lighting, thumbs-down when both of the user's hands are lifted and thumbs-down when the user is highly occluded.}
    \label{fig:edgefailure}
\end{figure*}

Examples of failure in gesture recognition are given in this section. Figure \ref{fig:failure} demonstrates a few general fails in gesture recognition. In these cases, it is assumed that HQ-Net did not manage to improve the quality of the image, thus, GViT provided the wrong recognition. However, the occurrence of these failed predictions happens with a very low probability while GViT reaches a 98.1\% success rate. Similarly, failures in several edge cases are given in Figure \ref{fig:edgefailure}. Nevertheless, in all cases, the model's certainties regarding its predictions are low. Hence, the framework can keep retaking images until acquiring a high certainty recognition.


\section*{Declaration of competing interest}

The authors declare that they have no known competing financial interests or personal relationships that could have appeared to influence the work reported in this paper.

\section*{Funding}

This work was supported by the Israel Innovation Authority (grant No. 77857).


\bibliographystyle{elsarticle-num} 
\bibliography{ref}

\begin{thebibliography}{10}
\expandafter\ifx\csname url\endcsname\relax
  \def\url#1{\texttt{#1}}\fi
\expandafter\ifx\csname urlprefix\endcsname\relax\def\urlprefix{URL }\fi
\expandafter\ifx\csname href\endcsname\relax
  \def\href#1#2{#2} \def\path#1{#1}\fi

\bibitem{krauss1996nonverbal}
R.~M. Krauss, Y.~Chen, P.~Chawla, Nonverbal behavior and nonverbal
  communication: What do conversational hand gestures tell us?, in: Advances in
  experimental social psychology, Vol.~28, Elsevier, 1996, pp. 389--450.

\bibitem{bernardis2006speech}
P.~Bernardis, M.~Gentilucci, Speech and gesture share the same communication
  system, Neuropsychologia 44~(2) (2006) 178--190.

\bibitem{cook2018enhancing}
S.~W. Cook, Enhancing learning with hand gestures: Potential mechanisms, in:
  Psychology of Learning and Motivation, Vol.~69, Elsevier, 2018, pp. 107--133.

\bibitem{goldin1999role}
S.~Goldin-Meadow, The role of gesture in communication and thinking, Trends in
  cognitive sciences 3~(11) (1999) 419--429.

\bibitem{bamani2023recognition}
E.~Bamani, E.~Nissinman, L.~Koenigsberg, I.~Meir, Y.~Matalon, A.~Sintov,
  Recognition and estimation of human finger pointing with an {RGB} camera for
  robot directive, arXiv preprint arXiv:2307.02949 (2023).

\bibitem{nickel2007visual}
K.~Nickel, R.~Stiefelhagen, Visual recognition of pointing gestures for
  human--robot interaction, Image and vision computing 25~(12) (2007)
  1875--1884.

\bibitem{wachs2011vision}
J.~P. Wachs, M.~K{\"o}lsch, H.~Stern, Y.~Edan, Vision-based hand-gesture
  applications, Communications of the ACM 54~(2) (2011) 60--71.

\bibitem{xia2019vision}
Z.~Xia, Q.~Lei, Y.~Yang, H.~Zhang, Y.~He, W.~Wang, M.~Huang, Vision-based hand
  gesture recognition for human-robot collaboration: a survey, in:
  International Conference on Control, Automation and Robotics (ICCAR), 2019,
  pp. 198--205.

\bibitem{weng2022development}
W.-T. Weng, H.-P. Huang, Y.-L. Zhao, C.-Y. Lin, Development of a visual
  perception system on a dual-arm mobile robot for human-robot interaction,
  Sensors 22~(23) (2022) 9545.

\bibitem{Gao2022}
Q.~Gao, Y.~Chen, Z.~Ju, Y.~Liang, Dynamic hand gesture recognition based on
  {3D} hand pose estimation for human–robot interaction, IEEE Sensors Journal
  22~(18) (2022) 17421--17430.
\newblock \href {https://doi.org/10.1109/JSEN.2021.3059685}
  {\path{doi:10.1109/JSEN.2021.3059685}}.

\bibitem{Liu2022}
D.~Liu, L.~Zhang, Y.~Wu, {LD-ConGR}: A large {RGB-D} video dataset for
  long-distance continuous gesture recognition, in: IEEE/CVF Conference on
  Computer Vision and Pattern Recognition (CVPR), 2022, pp. 3294--3302.
\newblock \href {https://doi.org/10.1109/CVPR52688.2022.00330}
  {\path{doi:10.1109/CVPR52688.2022.00330}}.

\bibitem{oudah2020hand}
M.~Oudah, A.~Al-Naji, J.~Chahl, Hand gesture recognition based on computer
  vision: a review of techniques, journal of Imaging 6~(8) (2020) 73.

\bibitem{An2022}
S.~An, X.~Zhang, D.~Wei, H.~Zhu, J.~Yang, K.~A. Tsintotas, {FastHand}: Fast
  monocular hand pose estimation on embedded systems, Journal of Systems
  Architecture 122 (2022) 102361.

\bibitem{AlbaCastro2014}
J.~L. Alba-Castro, E.~González~Agulla, F.~Loira, Hand gestures to control
  infotainment equipment in cars, IEEE Intelligent Vehicles Symposium,
  Proceedings (06 2014).
\newblock \href {https://doi.org/10.1109/IVS.2014.6856614}
  {\path{doi:10.1109/IVS.2014.6856614}}.

\bibitem{Buddhikot2018}
A.~G. Buddhikot, P.~Commerce, N.~M. Kulkarni, A.~Shaligram, Hand gesture
  interface based on skin detection technique for automotive infotainment
  system, International Journal of Image, Graphics and Signal Processing 10
  (2018) 10--24.

\bibitem{Deller2006}
M.~Deller, A.~Ebert, M.~Bender, H.~Hagen, Flexible gesture recognition for
  immersive virtual environments, in: International Conference on Information
  Visualisation, 2006, pp. 563--568.
\newblock \href {https://doi.org/10.1109/IV.2006.55}
  {\path{doi:10.1109/IV.2006.55}}.

\bibitem{Zulpukharkyzy2021}
L.~Zulpukharkyzy~Zholshiyeva, T.~Kokenovna~Zhukabayeva, S.~Turaev,
  M.~Aimambetovna~Berdiyeva, D.~Tokhtasynovna~Jambulova, Hand gesture
  recognition methods and applications: A literature survey, in: International
  Conference on Engineering \& MIS, Association for Computing Machinery, New
  York, NY, USA, 2021.
\newblock \href {https://doi.org/10.1145/3492547.3492578}
  {\path{doi:10.1145/3492547.3492578}}.

\bibitem{Mazhar2018}
O.~Mazhar, S.~Ramdani, B.~Navarro, R.~Passama, A.~Cherubini, Towards real-time
  physical human-robot interaction using skeleton information and hand
  gestures, in: IEEE/RSJ International Conference on Intelligent Robots and
  Systems (IROS), 2018, pp. 1--6.
\newblock \href {https://doi.org/10.1109/IROS.2018.8594385}
  {\path{doi:10.1109/IROS.2018.8594385}}.

\bibitem{Chang2019}
J.-Y. Chang, A.~Tejero-de Pablos, T.~Harada, Improved optical flow for
  gesture-based human-robot interaction, in: International Conference on
  Robotics and Automation (ICRA), 2019, pp. 7983--7989.
\newblock \href {https://doi.org/10.1109/ICRA.2019.8793825}
  {\path{doi:10.1109/ICRA.2019.8793825}}.

\bibitem{Iengo2014}
S.~Iengo, S.~Rossi, M.~Staffa, A.~Finzi, Continuous gesture recognition for
  flexible human-robot interaction, in: IEEE International Conference on
  Robotics and Automation (ICRA), 2014, pp. 4863--4868.
\newblock \href {https://doi.org/10.1109/ICRA.2014.6907571}
  {\path{doi:10.1109/ICRA.2014.6907571}}.

\bibitem{Nguyen2019}
X.~Nguyen, L.~Brun, O.~Lezoray, S.~Bougleux, A neural network based on {SPD}
  manifold learning for skeleton-based hand gesture recognition, in: IEEE/CVF
  Conference on Computer Vision and Pattern Recognition (CVPR), 2019, pp.
  12028--12037.

\bibitem{Jiang2021}
S.~Jiang, B.~Sun, L.~Wang, Y.~Bai, K.~Li, Y.~Fu, Skeleton aware multi-modal
  sign language recognition, in: IEEE/CVF Conference on Computer Vision and
  Pattern Recognition (CVPR) Workshops, 2021.

\bibitem{Cao2019}
Z.~{Cao}, G.~{Hidalgo Martinez}, T.~{Simon}, S.~{Wei}, Y.~A. {Sheikh},
  {OpenPose}: Realtime multi-person {2D} pose estimation using part affinity
  fields, IEEE Transactions on Pattern Analysis and Machine Intelligence
  (2019).

\bibitem{zhou2021long}
L.~Zhou, C.~Du, Z.~Sun, T.~L. Lam, Y.~Xu, Long-range hand gesture recognition
  via attention-based {SSD} network, in: IEEE International Conference on
  Robotics and Automation (ICRA), 2021, pp. 1832--1838.

\bibitem{Liang2024}
H.~Liang, L.~Fei, S.~Zhao, J.~Wen, S.~Teng, Y.~Xu, Mask-guided multiscale
  feature aggregation network for hand gesture recognition, Pattern Recognition
  145 (2024) 109901.
\newblock \href {https://doi.org/https://doi.org/10.1016/j.patcog.2023.109901}
  {\path{doi:https://doi.org/10.1016/j.patcog.2023.109901}}.

\bibitem{WangZ2021}
Z.~Wang, J.~Chen, S.~C.~H. Hoi, Deep learning for image super-resolution: A
  survey, IEEE Transactions on Pattern Analysis and Machine Intelligence
  43~(10) (2021) 3365--3387.
\newblock \href {https://doi.org/10.1109/TPAMI.2020.2982166}
  {\path{doi:10.1109/TPAMI.2020.2982166}}.

\bibitem{Wang2019}
X.~Wang, K.~Yu, S.~Wu, J.~Gu, Y.~Liu, C.~Dong, Y.~Qiao, C.~C. Loy, {ESRGAN}:
  Enhanced super-resolution generative adversarial networks, in:
  L.~Leal-Taix{\'e}, S.~Roth (Eds.), Computer Vision -- ECCV 2018 Workshops,
  Springer International Publishing, 2019, pp. 63--79.

\bibitem{Wang2021}
X.~Wang, L.~Xie, C.~Dong, Y.~Shan, Real-{ESRGAN}: Training real-world blind
  super-resolution with pure synthetic data, IEEE/CVF International Conference
  on Computer Vision Workshops (ICCVW) (2021) 1905--1914.

\bibitem{Ullah2019}
I.~Ullah, M.~Manzo, M.~Shah, M.~G. Madden, Graph convolutional networks:
  analysis, improvements and results, Applied Intelligence 52 (2019) 9033 --
  9044.

\bibitem{dosovitskiy2020image}
A.~Dosovitskiy, L.~Beyer, A.~Kolesnikov, D.~Weissenborn, X.~Zhai,
  T.~Unterthiner, M.~Dehghani, M.~Minderer, G.~Heigold, S.~Gelly, J.~Uszkoreit,
  N.~Houlsby, An image is worth 16x16 words: {Transformers} for image
  recognition at scale, in: International Conference on Learning
  Representations, 2021.

\bibitem{Brethes2004}
L.~Brethes, P.~Menezes, F.~Lerasle, J.~Hayet, Face tracking and hand gesture
  recognition for human-robot interaction, in: IEEE International Conference on
  Robotics and Automation, Vol.~2, 2004, pp. 1901--1906 Vol.2.
\newblock \href {https://doi.org/10.1109/ROBOT.2004.1308101}
  {\path{doi:10.1109/ROBOT.2004.1308101}}.

\bibitem{ma2018kinect}
X.~Ma, J.~Peng, Kinect sensor-based long-distance hand gesture recognition and
  fingertip detection with depth information, Journal of Sensors (2018) 1--9.

\bibitem{zhu2017multimodal}
G.~Zhu, L.~Zhang, P.~Shen, J.~Song, Multimodal gesture recognition using {3-D}
  convolution and convolutional {LSTM}, IEEE Access 5 (2017) 4517--4524.

\bibitem{nakamura2023deepoint}
S.~Nakamura, Y.~Kawanishi, S.~Nobuhara, K.~Nishino, {DeePoint}: Pointing
  recognition and direction estimation from a fixed view, in: IEEE/CVF
  International Conference on Computer Vision (ICCV), 2023.

\bibitem{Jirak2020}
D.~Jirak, D.~Biertimpel, M.~Kerzel, S.~Wermter, Solving visual object
  ambiguities when pointing: an unsupervised learning approach, Neural
  Computing and Applications (2020) 1--23.

\bibitem{Huang2009}
D.-Y. Huang, W.-C. Hu, S.-H. Chang, Vision-based hand gesture recognition using
  {PCA+Gabor} filters and {SVM}, in: International Conference on Intelligent
  Information Hiding and Multimedia Signal Processing, 2009, pp. 1--4.
\newblock \href {https://doi.org/10.1109/IIH-MSP.2009.96}
  {\path{doi:10.1109/IIH-MSP.2009.96}}.

\bibitem{Ziaie2009}
P.~Ziaie, T.~M{\"u}ller, M.~E. Foster, A.~Knoll, A {Na}{\"i}ve {Bayes}
  classifier with distance weighting for hand-gesture recognition, in:
  H.~Sarbazi-Azad, B.~Parhami, S.-G. Miremadi, S.~Hessabi (Eds.), Advances in
  Computer Science and Engineering, Springer Berlin Heidelberg, Berlin,
  Heidelberg, 2009, pp. 308--315.

\bibitem{eleni2015}
E.~Tsironi, P.~Barros, C.~Weber, S.~Wermter, An analysis of convolutional long
  short-term memory recurrent neural networks for gesture recognition,
  Neurocomputing 268 (2017) 76--86, advances in artificial neural networks,
  machine learning and computational intelligence.
\newblock \href {https://doi.org/https://doi.org/10.1016/j.neucom.2016.12.088}
  {\path{doi:https://doi.org/10.1016/j.neucom.2016.12.088}}.

\bibitem{kim2013vision}
D.~Kim, J.~Lee, H.-S. Yoon, J.~Kim, J.~Sohn, Vision-based arm gesture
  recognition for a long-range human--robot interaction, The Journal of
  Supercomputing 65 (2013) 336--352.

\bibitem{AlHammadi2020}
M.~Al-Hammadi, G.~Muhammad, W.~Abdul, M.~Alsulaiman, M.~A. Bencherif, M.~A.
  Mekhtiche, Hand gesture recognition for sign language using {3DCNN}, IEEE
  Access 8 (2020) 79491--79509.
\newblock \href {https://doi.org/10.1109/ACCESS.2020.2990434}
  {\path{doi:10.1109/ACCESS.2020.2990434}}.

\bibitem{lai20163d}
Y.~Lai, C.~Wang, Y.~Li, S.~S. Ge, D.~Huang, {3D} pointing gesture recognition
  for human-robot interaction, in: Chinese Control and Decision Conference
  (CCDC), IEEE, 2016, pp. 4959--4964.

\bibitem{fu2019research}
Y.~Fu, L.~Miao, Z.~Li, Research on long-distance hand recognition based on
  depth information, in: Journal of Physics: Conference Series, Vol. 1187, IOP
  Publishing, 2019, p. 042108.

\bibitem{mediapipe2019}
C.~Lugaresi, J.~Tang, H.~Nash, C.~McClanahan, E.~Uboweja, M.~Hays, F.~Zhang,
  C.-L. Chang, M.~Yong, J.~Lee, W.-T. Chang, W.~Hua, M.~Georg, M.~Grundmann,
  {MediaPipe}: A framework for perceiving and processing reality, in: Third
  Workshop on Computer Vision for AR/VR at IEEE Computer Vision and Pattern
  Recognition (CVPR) 2019, 2019.

\bibitem{Qiao2017}
S.~Qiao, Y.~Wang, J.~Li, Real-time human gesture grading based on openpose, in:
  International Congress on Image and Signal Processing, BioMedical Engineering
  and Informatics (CISP-BMEI), 2017, pp. 1--6.
\newblock \href {https://doi.org/10.1109/CISP-BMEI.2017.8301910}
  {\path{doi:10.1109/CISP-BMEI.2017.8301910}}.

\bibitem{Aiman2024}
U.~Aiman, T.~Ahmad, Angle based hand gesture recognition using graph
  convolutional network, Computer Animation and Virtual Worlds 35~(1) (2024)
  e2207.

\bibitem{Li2019gcn}
Y.~Li, Z.~He, X.~Ye, Z.~He, K.~Han, Spatial temporal graph convolutional
  networks for skeleton-based dynamic hand gesture recognition, J. Image Video
  Process. 2019~(1) (2019) 1–7.

\bibitem{Fang2020}
Z.~Fang, W.~Zhang, Z.~Guo, R.~Zhi, B.~Wang, F.~Flohr, Traffic police gesture
  recognition by pose graph convolutional networks, in: IEEE Intelligent
  Vehicles Symposium (IV), 2020, pp. 1833--1838.

\bibitem{Tchantchane2023}
R.~Tchantchane, H.~Zhou, S.~Zhang, G.~Alici, A review of hand gesture
  recognition systems based on noninvasive wearable sensors, Advanced
  Intelligent Systems 5~(10) (2023) 2300207.
\newblock \href {https://doi.org/https://doi.org/10.1002/aisy.202300207}
  {\path{doi:https://doi.org/10.1002/aisy.202300207}}.

\bibitem{Wang2023}
H.~Wang, B.~Ru, X.~Miao, Q.~Gao, M.~Habib, L.~Liu, S.~Qiu, {MEMS} devices-based
  hand gesture recognition via wearable computing, Micromachines 14~(5) (2023).
\newblock \href {https://doi.org/10.3390/mi14050947}
  {\path{doi:10.3390/mi14050947}}.

\bibitem{Alemayoh2022}
T.~T. Alemayoh, M.~Shintani, J.~H. Lee, S.~Okamoto, Deep-learning-based
  character recognition from handwriting motion data captured using {IMU} and
  force sensors, Sensors 22~(20) (2022).
\newblock \href {https://doi.org/10.3390/s22207840}
  {\path{doi:10.3390/s22207840}}.

\bibitem{Bongiovanni2023}
A.~Bongiovanni, A.~De~Luca, L.~Gava, L.~Grassi, M.~Lagomarsino, M.~Lapolla,
  A.~Marino, P.~Roncagliolo, S.~Macci{\`o}, A.~Carf{\`i}, F.~Mastrogiovanni,
  Gestural and touchscreen interaction for human-robot collaboration: A
  comparative study, in: I.~Petrovic, E.~Menegatti, I.~Markovi{\'{c}} (Eds.),
  Intelligent Autonomous Systems 17, Springer Nature Switzerland, Cham, 2023,
  pp. 122--138.

\bibitem{Benalcazar2017}
M.~E. Benalcázar, C.~Motoche, J.~A. Zea, A.~G. Jaramillo, C.~E. Anchundia,
  P.~Zambrano, M.~Segura, F.~Benalcázar~Palacios, M.~Pérez, Real-time hand
  gesture recognition using the {Myo} armband and muscle activity detection,
  in: IEEE Second Ecuador Technical Chapters Meeting (ETCM), 2017, pp. 1--6.
\newblock \href {https://doi.org/10.1109/ETCM.2017.8247458}
  {\path{doi:10.1109/ETCM.2017.8247458}}.

\bibitem{Moin2020}
A.~Moin, A.~Zhou, A.~Rahimi, A.~Menon, S.~Benatti, G.~Alexandrov, S.~Tamakloe,
  J.~Ting, N.~Yamamoto, Y.~Khan, F.~L. Burghardt, L.~Benini, A.~C. Arias, J.~M.
  Rabaey, A wearable biosensing system with in-sensor adaptive machine learning
  for hand gesture recognition, Nature Electronics 4 (2020) 54--63.

\bibitem{Lian2017}
K.-Y. Lian, C.-C. Chiu, Y.-J. Hong, W.-T. Sung, Wearable armband for real time
  hand gesture recognition, in: IEEE International Conference on Systems, Man,
  and Cybernetics (SMC), 2017, pp. 2992--2995.
\newblock \href {https://doi.org/10.1109/SMC.2017.8123083}
  {\path{doi:10.1109/SMC.2017.8123083}}.

\bibitem{Kim2023}
E.~Kim, J.~Shin, Y.~Kwon, B.~Park, {EMG}-based dynamic hand gesture recognition
  using edge {AI} for human–robot interaction, Electronics 12~(7) (2023)
  1541.
\newblock \href {https://doi.org/10.3390/electronics12071541}
  {\path{doi:10.3390/electronics12071541}}.

\bibitem{Fora2021}
M.~Fora, B.~Ben~Atitallah, K.~Lweesy, O.~Kanoun, Hand gesture recognition based
  on force myography measurements using {KNN} classifier, in: International
  Multi-Conference on Systems, Signals \& Devices (SSD), 2021, pp. 960--964.
\newblock \href {https://doi.org/10.1109/SSD52085.2021.9429514}
  {\path{doi:10.1109/SSD52085.2021.9429514}}.

\bibitem{Negreiros2022}
M.~N. Rylo, R.~L. {de Medeiros}, V.~F. {de Lucena Jr}, Gesture recognition of
  wrist motion based on wearables sensors, Procedia Computer Science 210 (2022)
  181--188, international Conference on Emerging Ubiquitous Systems and
  Pervasive Networks (EUSPN).
\newblock \href {https://doi.org/https://doi.org/10.1016/j.procs.2022.10.135}
  {\path{doi:https://doi.org/10.1016/j.procs.2022.10.135}}.

\bibitem{Siddiqui2017}
N.~Siddiqui, R.~H.~M. Chan, A wearable hand gesture recognition device based on
  acoustic measurements at wrist, in: Annual International Conference of the
  IEEE Engineering in Medicine and Biology Society (EMBC), 2017, pp.
  4443--4446.
\newblock \href {https://doi.org/10.1109/EMBC.2017.8037842}
  {\path{doi:10.1109/EMBC.2017.8037842}}.

\bibitem{Bandini2023}
A.~Bandini, J.~Zariffa, Analysis of the hands in egocentric vision: A survey,
  IEEE Trans. Pattern Anal. Mach. Intell. 45~(6) (2023) 6846–6866.

\bibitem{Alam2022}
M.~M. Alam, M.~T. Islam, S.~M. Rahman, Unified learning approach for egocentric
  hand gesture recognition and fingertip detection, Pattern Recognition 121
  (2022) 108200.

\bibitem{Sundaramoorthy2021}
S.~Sundaramoorthy, B.~Muthazhagan, Super-Resolution-Based Human-Computer
  Interaction System for Speech and Hearing Impaired Using Real-Time Hand
  Gesture Recognition System, Springer International Publishing, Cham, 2021,
  pp. 135--153.

\bibitem{Li2020}
Y.~Li, G.~Dong, P.~Huang, Z.~Ma, X.~Wang, A gesture recognition framework based
  on multi-frame super-resolution image sequence, in: Chinese Automation
  Congress (CAC), 2020, pp. 4519--4524.

\bibitem{Kaur2021}
J.~Kaur, N.~Mittal, S.~Kaur, Hand gesture image enhancement for improved
  recognition and subsequent analysis, in: R.~S. Tomar, S.~Verma, B.~K.
  Chaurasia, V.~Singh, J.~Abawajy, S.~Akashe, P.-A. Hsiung, V.~K. Bhargava
  (Eds.), Communication, Networks and Computing, Springer Singapore, 2021, pp.
  354--365.

\bibitem{Nasrollahi2014}
K.~Nasrollahi, T.~B. Moeslund, Super-resolution: a comprehensive survey,
  Machine Vision and Applications 25 (2014) 1423 -- 1468.

\bibitem{Anwar2020}
S.~Anwar, S.~Khan, N.~Barnes, A deep journey into super-resolution: A survey,
  ACM Comput. Surv. 53~(3) (2020).

\bibitem{Hu2019}
X.~Hu, M.~A. Naiel, A.~Wong, M.~Lamm, P.~Fieguth, {RUNet}: A robust {UNet}
  architecture for image super-resolution, in: IEEE/CVF Conference on Computer
  Vision and Pattern Recognition Workshops (CVPRW), 2019, pp. 505--507.
\newblock \href {https://doi.org/10.1109/CVPRW.2019.00073}
  {\path{doi:10.1109/CVPRW.2019.00073}}.

\bibitem{ronneberger2015}
O.~Ronneberger, P.~Fischer, T.~Brox, {U-Net}: Convolutional networks for
  biomedical image segmentation, in: Medical Image Computing and
  Computer-Assisted Intervention, 2015, pp. 234--241.

\bibitem{Lu2019}
Z.~Lu, Y.~Chen, Single image super-resolution based on a modified {U-Net} with
  mixed gradient loss, Signal, Image and Video Processing 16 (2019) 1143 --
  1151.

\bibitem{zhang2021}
K.~Zhang, J.~Liang, L.~Van~Gool, R.~Timofte, Designing a practical degradation
  model for deep blind image super-resolution, in: IEEE International
  Conference on Computer Vision, 2021, pp. 4791--4800.

\bibitem{WangZhihao2019}
Z.~Wang, J.~Chen, S.~C.~H. Hoi,
  \href{https://api.semanticscholar.org/CorpusID:62841491}{Deep learning for
  image super-resolution: A survey}, IEEE Transactions on Pattern Analysis and
  Machine Intelligence 43 (2019) 3365--3387.
\newline\urlprefix\url{https://api.semanticscholar.org/CorpusID:62841491}

\bibitem{Ye2023}
S.~Ye, S.~Zhao, Y.~Hu, C.~Xie,
  \href{https://www.mdpi.com/2079-9292/12/13/2975}{Single-image
  super-resolution challenges: A brief review}, Electronics 12~(13) (2023).
\newline\urlprefix\url{https://www.mdpi.com/2079-9292/12/13/2975}

\bibitem{redmon2018yolov3}
J.~Redmon, A.~Farhadi, {YOLOv3}: An incremental improvement, arXiv preprint
  arXiv:1804.02767 (2018).

\bibitem{Li2019}
G.~Li, M.~Muller, A.~Thabet, B.~Ghanem,
  \href{https://doi.ieeecomputersociety.org/10.1109/ICCV.2019.00936}{{DeepGCNs}:
  Can {GCN}s go as deep as {CNN}s?}, in: IEEE/CVF International Conference on
  Computer Vision (ICCV), IEEE Computer Society, 2019, pp. 9266--9275.
\newblock \href {https://doi.org/10.1109/ICCV.2019.00936}
  {\path{doi:10.1109/ICCV.2019.00936}}.
\newline\urlprefix\url{https://doi.ieeecomputersociety.org/10.1109/ICCV.2019.00936}

\bibitem{Vasudevan2023}
V.~Vasudevan, M.~Bassenne, M.~T. Islam, L.~Xing,
  \href{https://doi.org/10.1016/j.patrec.2023.01.003}{Image classification
  using graph neural network and multiscale wavelet superpixels}, Pattern
  Recogn. Lett. 166~(C) (2023) 89–96.
\newline\urlprefix\url{https://doi.org/10.1016/j.patrec.2023.01.003}

\bibitem{Vaswani2017}
A.~Vaswani, N.~Shazeer, N.~Parmar, J.~Uszkoreit, L.~Jones, A.~N. Gomez,
  L.~Kaiser, I.~Polosukhin, Attention is all you need, in: International
  Conference on Neural Information Processing Systems, 2017, p. 6000–6010.

\bibitem{Bae2022}
J.-H. Bae, G.-H. Yu, J.-H. Lee, D.~T. Vu, L.~H. Anh, H.-G. Kim, J.-Y. Kim,
  \href{https://www.mdpi.com/2076-3417/12/18/9176}{Superpixel image
  classification with graph convolutional neural networks based on learnable
  positional embedding}, Applied Sciences 12~(18) (2022).
\newline\urlprefix\url{https://www.mdpi.com/2076-3417/12/18/9176}

\bibitem{Dauphin2017}
Y.~N. Dauphin, A.~Fan, M.~Auli, D.~Grangier, Language modeling with gated
  convolutional networks, in: International Conference on Machine Learning,
  Vol.~70, 2017, p. 933–941.

\bibitem{liaw2018tune}
R.~Liaw, E.~Liang, R.~Nishihara, P.~Moritz, J.~E. Gonzalez, I.~Stoica, Tune: A
  research platform for distributed model selection and training, arXiv
  preprint arXiv:1807.05118 (2018).

\bibitem{Zeng2017}
K.~Zeng, J.~Yu, R.~Wang, C.~Li, D.~Tao, Coupled deep autoencoder for single
  image super-resolution, IEEE Transactions on Cybernetics 47~(1) (2017)
  27--37.
\newblock \href {https://doi.org/10.1109/TCYB.2015.2501373}
  {\path{doi:10.1109/TCYB.2015.2501373}}.

\bibitem{zhou2019unet++}
Z.~Zhou, M.~M.~R. Siddiquee, N.~Tajbakhsh, J.~Liang, {UNet++}: Redesigning skip
  connections to exploit multiscale features in image segmentation, IEEE
  transactions on medical imaging 39~(6) (2019) 1856--1867.

\bibitem{Chen2022}
H.~Chen, X.~He, L.~Qing, Y.~Wu, C.~Ren, R.~E. Sheriff, C.~Zhu, Real-world
  single image super-resolution: A brief review, Information Fusion 79 (2022)
  124--145.
\newblock \href {https://doi.org/https://doi.org/10.1016/j.inffus.2021.09.005}
  {\path{doi:https://doi.org/10.1016/j.inffus.2021.09.005}}.

\bibitem{huang2017densely}
G.~Huang, Z.~Liu, L.~Van Der~Maaten, K.~Q. Weinberger, Densely connected
  convolutional networks, in: IEEE Conference on Computer Vision and Pattern
  Recognition, 2017, pp. 4700--4708.

\bibitem{tan2019}
M.~Tan, Q.~Le, Efficientnet: Rethinking model scaling for convolutional neural
  networks, in: International conference on machine learning, 2019, pp.
  6105--6114.

\bibitem{szegedy2015going}
C.~Szegedy, W.~Liu, Y.~Jia, P.~Sermanet, S.~Reed, D.~Anguelov, D.~Erhan,
  V.~Vanhoucke, A.~Rabinovich, Going deeper with convolutions, in: IEEE
  conference on computer vision and pattern recognition, 2015, pp. 1--9.

\bibitem{Zagoruyko2016}
S.~Zagoruyko, N.~Komodakis, Wide residual networks, in: British Machine Vision
  Conference {BMVC}, 2016.

\bibitem{simonyan2014very}
K.~Simonyan, A.~Zisserman, Very deep convolutional networks for large-scale
  image recognition, in: International Conference on Learning Representations
  (ICLR 2015), Computational and Biological Learning Society, 2015, pp. 1--14.

\end{thebibliography}

\end{document}